\documentclass[12pt]{article}

\usepackage{etex}

\usepackage{tikz}
\usetikzlibrary{cd,decorations,shapes}
\newenvironment{tikzar}[1][]{{}\kern-4pt\begin{tikzcd}[ampersand replacement=\&,#1]}%
{\end{tikzcd}\kern-4pt{}}

\usepackage{graphicx}
\usepackage{pstricks}

\usepackage{bbold}
\usepackage{bbm}
\usepackage%[only,sslash]
{stmaryrd}

\usepackage[USenglish]{babel}
\usepackage{amssymb}
 \usepackage{amstext}
 \usepackage{amsmath}
\usepackage{amsthm}

\usepackage{enumerate}

\newdimen\proofrulebreadth \proofrulebreadth=.05em
\newdimen\proofdotseparation \proofdotseparation=1.25ex
\newdimen\proofrulebaseline \proofrulebaseline=2ex
\newcount\proofdotnumber \proofdotnumber=3
\let\then\relax
\def\hfi{\hskip0pt plus.0001fil}
\mathchardef\squigto="3A3B
\newif\ifinsideprooftree\insideprooftreefalse
\newif\ifonleftofproofrule\onleftofproofrulefalse
\newif\ifproofdots\proofdotsfalse
\newif\ifdoubleproof\doubleprooffalse
\let\wereinproofbit\relax
\newdimen\shortenproofleft
\newdimen\shortenproofright
\newdimen\proofbelowshift
\newbox\proofabove
\newbox\proofbelow
\newbox\proofrulename
\def\shiftproofbelow{\let\next\relax\afterassignment\setshiftproofbelow\dimen0 }
\def\shiftproofbelowneg{\def\next{\multiply\dimen0 by-1 }%
\afterassignment\setshiftproofbelow\dimen0 }
\def\setshiftproofbelow{\next\proofbelowshift=\dimen0 }
\def\setproofrulebreadth{\proofrulebreadth}

\def\prooftree{% NESTED ZERO (\ifonleftofproofrule)
\ifnum  \lastpenalty=1
\then   \unpenalty
\else   \onleftofproofrulefalse
\fi
\ifonleftofproofrule
\else   \ifinsideprooftree
        \then   \hskip.5em plus1fil
        \fi
\fi
\bgroup% NESTED ONE (\proofbelow, \proofrulename, \proofabove,
\setbox\proofbelow=\hbox{}\setbox\proofrulename=\hbox{}%
\let\justifies\proofover\let\leadsto\proofoverdots\let\Justifies\proofoverdbl
\let\using\proofusing\let\[\prooftree
\ifinsideprooftree\let\]\endprooftree\fi
\proofdotsfalse\doubleprooffalse
\let\thickness\setproofrulebreadth
\let\shiftright\shiftproofbelow \let\shift\shiftproofbelow
\let\shiftleft\shiftproofbelowneg
\let\ifwasinsideprooftree\ifinsideprooftree
\insideprooftreetrue
\setbox\proofabove=\hbox\bgroup$\displaystyle % NESTED TWO
\let\wereinproofbit\prooftree
\shortenproofleft=0pt \shortenproofright=0pt \proofbelowshift=0pt
\onleftofproofruletrue\penalty1
}

\def\eproofbit{% NESTED TWO
\ifx    \wereinproofbit\prooftree
\then   \ifcase \lastpenalty
        \then   \shortenproofright=0pt  % 0: some other object, no indentation
        \or     \unpenalty\hfil         % 1: empty hypotheses, just glue
        \or     \unpenalty\unskip       % 2: just had a tree, remove glue
        \else   \shortenproofright=0pt  % eh?
        \fi
\fi
\global\dimen0=\shortenproofleft
\global\dimen1=\shortenproofright
\global\dimen2=\proofrulebreadth
\global\dimen3=\proofbelowshift
\global\dimen4=\proofdotseparation
\global\count255=\proofdotnumber
$\egroup  % NESTED ONE
\shortenproofleft=\dimen0
\shortenproofright=\dimen1
\proofrulebreadth=\dimen2
\proofbelowshift=\dimen3
\proofdotseparation=\dimen4
\proofdotnumber=\count255
}

\def\proofover{% NESTED TWO
\eproofbit % NESTED ONE
\setbox\proofbelow=\hbox\bgroup % NESTED TWO
\let\wereinproofbit\proofover
$\displaystyle
}%
\def\proofoverdbl{% NESTED TWO
\eproofbit % NESTED ONE
\doubleprooftrue
\setbox\proofbelow=\hbox\bgroup % NESTED TWO
\let\wereinproofbit\proofoverdbl
$\displaystyle
}%
\def\proofoverdots{% NESTED TWO
\eproofbit % NESTED ONE
\proofdotstrue
\setbox\proofbelow=\hbox\bgroup % NESTED TWO
\let\wereinproofbit\proofoverdots
$\displaystyle
}%
\def\proofusing{% NESTED TWO
\eproofbit % NESTED ONE
\setbox\proofrulename=\hbox\bgroup % NESTED TWO
\let\wereinproofbit\proofusing
\kern0.3em$
}

\def\endprooftree{% NESTED TWO
\eproofbit % NESTED ONE
  \dimen5 =0pt% spread of hypotheses
\dimen0=\wd\proofabove \advance\dimen0-\shortenproofleft
\advance\dimen0-\shortenproofright
\dimen1=.5\dimen0 \advance\dimen1-.5\wd\proofbelow
\dimen4=\dimen1
\advance\dimen1\proofbelowshift \advance\dimen4-\proofbelowshift
\ifdim  \dimen1<0pt
\then   \advance\shortenproofleft\dimen1
        \advance\dimen0-\dimen1
        \dimen1=0pt
        \ifdim  \shortenproofleft<0pt
        \then   \setbox\proofabove=\hbox{%
                        \kern-\shortenproofleft\unhbox\proofabove}%
                \shortenproofleft=0pt
        \fi
\fi
\ifdim  \dimen4<0pt
\then   \advance\shortenproofright\dimen4
        \advance\dimen0-\dimen4
        \dimen4=0pt
\fi
\ifdim  \shortenproofright<\wd\proofrulename
\then   \shortenproofright=\wd\proofrulename
\fi
\dimen2=\shortenproofleft \advance\dimen2 by\dimen1
\dimen3=\shortenproofright\advance\dimen3 by\dimen4
\ifproofdots
\then
        \dimen6=\shortenproofleft \advance\dimen6 .5\dimen0
        \setbox1=\vbox to\proofdotseparation{\vss\hbox{$\cdot$}\vss}%
        \setbox0=\hbox{%
                \advance\dimen6-.5\wd1
                \kern\dimen6
                $\vcenter to\proofdotnumber\proofdotseparation
                        {\leaders\box1\vfill}$%
                \unhbox\proofrulename}%
\else   \dimen6=\fontdimen22\the\textfont2 % height of maths axis
        \dimen7=\dimen6
        \advance\dimen6by.5\proofrulebreadth
        \advance\dimen7by-.5\proofrulebreadth
        \setbox0=\hbox{%
                \kern\shortenproofleft
                \ifdoubleproof
                \then   \hbox to\dimen0{%
                        $\mathsurround0pt\mathord=\mkern-6mu%
                        \cleaders\hbox{$\mkern-2mu=\mkern-2mu$}\hfill
                        \mkern-6mu\mathord=$}%
                \else   \vrule height\dimen6 depth-\dimen7 width\dimen0
                \fi
                \unhbox\proofrulename}%
        \ht0=\dimen6 \dp0=-\dimen7
\fi
\let\doll\relax
\ifwasinsideprooftree
\then   \let\VBOX\vbox
\else   \ifmmode\else$\let\doll=$\fi
        \let\VBOX\vcenter
\fi
\VBOX   {\baselineskip\proofrulebaseline \lineskip.2ex
        \expandafter\lineskiplimit\ifproofdots0ex\else-0.6ex\fi
        \hbox   spread\dimen5   {\hfi\unhbox\proofabove\hfi}%
        \hbox{\box0}%
        \hbox   {\kern\dimen2 \box\proofbelow}}\doll%
\global\dimen2=\dimen2
\global\dimen3=\dimen3
\egroup % NESTED ZERO
\ifonleftofproofrule
\then   \shortenproofleft=\dimen2
\fi
\shortenproofright=\dimen3
\onleftofproofrulefalse
\ifinsideprooftree
\then   \hskip.5em plus 1fil \penalty2
\fi
}
\renewcommand{\to}{\xrightarrow{}}%{\longrightarrow}
\newcommand{\ot}{\xleftarrow{}}%{\longleftarrow}
\newcommand{\tto}[1]{\xrightarrow{#1}}
\newcommand{\oot}[1]{\xleftarrow{#1}}

\newcommand{\inclusion}{\hookrightarrow}

\newcommand{\Set}{\mathsf{Set}}

\newcommand{\id}{{\rm id}}

\newcommand{\EEE}{{\cal E}}

\newcommand{\MMM}{{\cal M}}

\newcommand{\OOO}{{\cal O}}

\newcommand{\TTT}{{\cal T}}
\newcommand{\UUU}{{\cal U}}

\renewcommand{\Bbb}{\mathbb}

\newcommand{\BBb}{{\Bbb B}}

\newcommand{\LLl}{{\Bbb L}}

\newcommand{\NNn}{{\Bbb N}}

\newcommand{\PPp}{{\Bbb P}}

\mathcode`\<="4268 %left delimiter
\mathcode`\>="5269 %right delimiter
\mathchardef\gt="313E %relation >
\mathchardef\lt="313C %relation <

\newcommand{\beq}{\begin{equation}}
\newcommand{\eeq}{\end{equation}}
\newcommand{\ba}[1]{\begin{array}{#1}}
\newcommand{\ea}{\end{array}}
\newcommand{\bea}{\begin{eqnarray}}
\newcommand{\eea}{\end{eqnarray}}
\newcommand{\bear}{\begin{eqnarray*}}
\newcommand{\eear}{\end{eqnarray*}}

\theoremstyle{plain}
\newtheorem{theorem}{Theorem}[section]

\newtheorem*{proposition*}{Proposition}

\newtheorem{proposition}[theorem]{Proposition}
\newtheorem{definition}[theorem]{Definition}

\newtheorem{lemma}[theorem]{Lemma}
\theoremstyle{remark}

\theoremstyle{remark}

\theoremstyle{remark}

\theoremstyle{remark}

\newenvironment{prf}[1]{\begin{trivlist} \item[{\bf ~Proof #1.}]}%
{\qed\end{trivlist}}

\newcommand{\bpr}{\begin{prf}{\!\!}}
\newcommand{\epr}{\end{prf}}
\newcommand{\bprf}[1]{\begin{prf}{#1}}
\newcommand{\eprf}{\end{prf}}

\usepackage{enumerate}
\usepackage{rotating}
\usepackage{amsfonts}
\usepackage{amssymb}
\usepackage{stmaryrd}
\usepackage{fullpage}
\usepackage{authblk}

\usepackage[LGR,T1]{fontenc}

\usepackage{hyperref}

\newdimen\proofrulebreadth \proofrulebreadth=.05em
\newdimen\proofdotseparation \proofdotseparation=1.25ex
\newdimen\proofrulebaseline \proofrulebaseline=2ex
\newcount\proofdotnumber \proofdotnumber=3
\let\then\relax
\def\hfi{\hskip0pt plus.0001fil}
\mathchardef\squigto="3A3B
\newif\ifinsideprooftree\insideprooftreefalse
\newif\ifonleftofproofrule\onleftofproofrulefalse
\newif\ifproofdots\proofdotsfalse
\newif\ifdoubleproof\doubleprooffalse
\let\wereinproofbit\relax
\newdimen\shortenproofleft
\newdimen\shortenproofright
\newdimen\proofbelowshift
\newbox\proofabove
\newbox\proofbelow
\newbox\proofrulename
\def\shiftproofbelow{\let\next\relax\afterassignment\setshiftproofbelow\dimen0 }
\def\shiftproofbelowneg{\def\next{\multiply\dimen0 by-1 }%
\afterassignment\setshiftproofbelow\dimen0 }
\def\setshiftproofbelow{\next\proofbelowshift=\dimen0 }
\def\setproofrulebreadth{\proofrulebreadth}

\def\prooftree{% NESTED ZERO (\ifonleftofproofrule)
\ifnum  \lastpenalty=1
\then   \unpenalty
\else   \onleftofproofrulefalse
\fi
\ifonleftofproofrule
\else   \ifinsideprooftree
        \then   \hskip.5em plus1fil
        \fi
\fi
\bgroup% NESTED ONE (\proofbelow, \proofrulename, \proofabove,
\setbox\proofbelow=\hbox{}\setbox\proofrulename=\hbox{}%
\let\justifies\proofover\let\leadsto\proofoverdots\let\Justifies\proofoverdbl
\let\using\proofusing\let\[\prooftree
\ifinsideprooftree\let\]\endprooftree\fi
\proofdotsfalse\doubleprooffalse
\let\thickness\setproofrulebreadth
\let\shiftright\shiftproofbelow \let\shift\shiftproofbelow
\let\shiftleft\shiftproofbelowneg
\let\ifwasinsideprooftree\ifinsideprooftree
\insideprooftreetrue
\setbox\proofabove=\hbox\bgroup$\displaystyle % NESTED TWO
\let\wereinproofbit\prooftree
\shortenproofleft=0pt \shortenproofright=0pt \proofbelowshift=0pt
\onleftofproofruletrue\penalty1
}

\def\eproofbit{% NESTED TWO
\ifx    \wereinproofbit\prooftree
\then   \ifcase \lastpenalty
        \then   \shortenproofright=0pt  % 0: some other object, no indentation
        \or     \unpenalty\hfil         % 1: empty hypotheses, just glue
        \or     \unpenalty\unskip       % 2: just had a tree, remove glue
        \else   \shortenproofright=0pt  % eh?
        \fi
\fi
\global\dimen0=\shortenproofleft
\global\dimen1=\shortenproofright
\global\dimen2=\proofrulebreadth
\global\dimen3=\proofbelowshift
\global\dimen4=\proofdotseparation
\global\count255=\proofdotnumber
$\egroup  % NESTED ONE
\shortenproofleft=\dimen0
\shortenproofright=\dimen1
\proofrulebreadth=\dimen2
\proofbelowshift=\dimen3
\proofdotseparation=\dimen4
\proofdotnumber=\count255
}

\def\proofover{% NESTED TWO
\eproofbit % NESTED ONE
\setbox\proofbelow=\hbox\bgroup % NESTED TWO
\let\wereinproofbit\proofover
$\displaystyle
}%
\def\proofoverdbl{% NESTED TWO
\eproofbit % NESTED ONE
\doubleprooftrue
\setbox\proofbelow=\hbox\bgroup % NESTED TWO
\let\wereinproofbit\proofoverdbl
$\displaystyle
}%
\def\proofoverdots{% NESTED TWO
\eproofbit % NESTED ONE
\proofdotstrue
\setbox\proofbelow=\hbox\bgroup % NESTED TWO
\let\wereinproofbit\proofoverdots
$\displaystyle
}%
\def\proofusing{% NESTED TWO
\eproofbit % NESTED ONE
\setbox\proofrulename=\hbox\bgroup % NESTED TWO
\let\wereinproofbit\proofusing
\kern0.3em$
}

\def\endprooftree{% NESTED TWO
\eproofbit % NESTED ONE
  \dimen5 =0pt% spread of hypotheses
\dimen0=\wd\proofabove \advance\dimen0-\shortenproofleft
\advance\dimen0-\shortenproofright
\dimen1=.5\dimen0 \advance\dimen1-.5\wd\proofbelow
\dimen4=\dimen1
\advance\dimen1\proofbelowshift \advance\dimen4-\proofbelowshift
\ifdim  \dimen1<0pt
\then   \advance\shortenproofleft\dimen1
        \advance\dimen0-\dimen1
        \dimen1=0pt
        \ifdim  \shortenproofleft<0pt
        \then   \setbox\proofabove=\hbox{%
                        \kern-\shortenproofleft\unhbox\proofabove}%
                \shortenproofleft=0pt
        \fi
\fi
\ifdim  \dimen4<0pt
\then   \advance\shortenproofright\dimen4
        \advance\dimen0-\dimen4
        \dimen4=0pt
\fi
\ifdim  \shortenproofright<\wd\proofrulename
\then   \shortenproofright=\wd\proofrulename
\fi
\dimen2=\shortenproofleft \advance\dimen2 by\dimen1
\dimen3=\shortenproofright\advance\dimen3 by\dimen4
\ifproofdots
\then
        \dimen6=\shortenproofleft \advance\dimen6 .5\dimen0
        \setbox1=\vbox to\proofdotseparation{\vss\hbox{$\cdot$}\vss}%
        \setbox0=\hbox{%
                \advance\dimen6-.5\wd1
                \kern\dimen6
                $\vcenter to\proofdotnumber\proofdotseparation
                        {\leaders\box1\vfill}$%
                \unhbox\proofrulename}%
\else   \dimen6=\fontdimen22\the\textfont2 % height of maths axis
        \dimen7=\dimen6
        \advance\dimen6by.5\proofrulebreadth
        \advance\dimen7by-.5\proofrulebreadth
        \setbox0=\hbox{%
                \kern\shortenproofleft
                \ifdoubleproof
                \then   \hbox to\dimen0{%
                        $\mathsurround0pt\mathord=\mkern-6mu%
                        \cleaders\hbox{$\mkern-2mu=\mkern-2mu$}\hfill
                        \mkern-6mu\mathord=$}%
                \else   \vrule height\dimen6 depth-\dimen7 width\dimen0
                \fi
                \unhbox\proofrulename}%
        \ht0=\dimen6 \dp0=-\dimen7
\fi
\let\doll\relax
\ifwasinsideprooftree
\then   \let\VBOX\vbox
\else   \ifmmode\else$\let\doll=$\fi
        \let\VBOX\vcenter
\fi
\VBOX   {\baselineskip\proofrulebaseline \lineskip.2ex
        \expandafter\lineskiplimit\ifproofdots0ex\else-0.6ex\fi
        \hbox   spread\dimen5   {\hfi\unhbox\proofabove\hfi}%
        \hbox{\box0}%
        \hbox   {\kern\dimen2 \box\proofbelow}}\doll%
\global\dimen2=\dimen2
\global\dimen3=\dimen3
\egroup % NESTED ZERO
\ifonleftofproofrule
\then   \shortenproofleft=\dimen2
\fi
\shortenproofright=\dimen3
\onleftofproofrulefalse
\ifinsideprooftree
\then   \hskip.5em plus 1fil \penalty2
\fi
}

\renewcommand{\paragraph}[1]{\vspace{.3\baselineskip}\noindent\textbf{#1}}

\renewcommand{\vec}[1]{\mathbf{#1}}
\newcommand{\DP}{\LLl}%{\PPp}

\newcommand{\sconvg}{\mbox{\scriptsize $\rotatebox[origin=c]{90}{$\multimap$}^{\raisebox{-.29ex}{\hspace{-1.2ex}$\scriptstyle\bullet$}}$}}

\newcommand{\cmn}{\Delta}%{\delta}%{\vartriangle}%
\newcommand{\scun}{\sconvg}

\newcommand{\tot}[1]{{#1}^\bullet}

\newcommand{\uev}[1]{\mbox{\large$\left\{\mbox{\normalsize $#1$}\right\}$}}
\newcommand{\pev}[1]{\left[{#1}\right]}%{\left\lceil{#1}\right\rceil}
\newcommand{\prtial}{\pev{\,}}
\newcommand{\universal}{\mbox{\large$\{\}$}}
\newcommand{\uuniversal}{\{\}}

\newcommand{\sta}[1]{{#1}^{\prime}}%{{#1}^{\triangleright}}%^\circ}
\newcommand{\out}[1]{{#1}^{\prime\prime}}%{{#1}^{\odot}}%{\triangleleft}}%\bullet}

\newcommand{\EQLS}{\mbox{\large$=$}}

\newcommand{\enco}[1]{\left\ulcorner{#1}\right\urcorner}%{\llceil{#1}\rrceil}%

\newcommand{\ana}[1]{\llbracket{#1}\rrbracket}

\newcommand{\Prov}{\P}%{{\sf P}}
\newcommand{\Bool}{\BBb}

\graphicspath{{PICS/}}

\begin{document}

\title{From G\"odel's Incompleteness Theorem\\
to the completeness of bot beliefs\\[.5ex]
{\Large (Extended abstract)}}

\author{Dusko~Pavlovic\thanks{Partially supported by NSF and AFOSR.}} 
\affil{University of Hawaii, Honolulu HI}
\author[**]{Temra Pavlovic}

\date{}

\maketitle

\begin{abstract} 
\noindent Hilbert and Ackermann asked for a method to consistently extend incomplete theories to complete theories. G\"odel essentially proved that any theory capable of encoding its own statements and their proofs contains statements that are true but not provable. Hilbert did not accept that G\"odel's construction answered his question, and in his late writings and lectures, G\"odel agreed that it did not, since theories can be completed incrementally, by adding axioms to prove ever more true statements, as science normally does, with completeness as the vanishing point. This pragmatic view of validity is familiar not only to scientists who conjecture test hypotheses but also to real-estate agents and other dealers, who conjure claims, albeit invalid, as necessary to close a deal, confident that they will be able to conjure other claims, albeit invalid, sufficient to make the first claims valid. We study the underlying logical process and describe the trajectories leading to testable but unfalsifiable theories to which bots and other automated learners are likely to converge.
\end{abstract}

\section{Introduction}% wrong, fake, artificial causation}
\label{Sec:Intro}
% !TEX root = 00-wollic.tex

%\input{10-wollic}
Logic as the theory of theories was originally developed to prove true statements. Here we study developments in the opposite direction: modifying interpretations to make true some previously false statements. In modal logic, such logical processes have been modeled as instances of \emph{belief update}\/ \cite{Baltag-Sadrzadeh-Coecke:DEL,Baltag-Moss:logic,Ditmarsch:DEL}. In the practice of science, such processes arise when theories are updated to explain new observations \cite[Ch.~4]{Osherson:sci-inquiry}. In public life, the goal of such processes is to influence some public perceptions to better suit some private preferences \cite[Part~V]{Easley-Kleinberg:book}. This range of applications gave rise to a gamut of techniques for  influence and belief engineering, covering the space from unsupervised learning to conditioning. 

\paragraph{From incomplete theories to complete beliefs.} The idea to incrementally complete incomplete theories \cite{DavisM:undecidable} arose soon after G\"odel proved his Incompleteness Theorem \cite{GoedelK:ueber}. Alan Turing wrote a  thesis about ordinal towers of completions and discovered the hierarchy of unsolvability degrees \cite{TuringA:thesis}. The core idea was to keep recognizing and adding true but unprovable statements to theories. In the meantime, interests shifted from making true statements provable to making false statements true. Many toy examples of belief updates and revisions have been formalized and studied in dynamic-epistemic logic \cite{Baltag-Sadrzadeh:DEL,BenthemJ:interaction}, but the advances in belief engineering and the resulting industry of influence overtook the theory at great speed, and turned several corners of market and political monetizations. The theory remained   fragmented even on its own. While modal presentations of G\"odel's theorems appeared early on \cite{RosserB:extensions}, the computational ideas, that made his self-referential constructions  possible \cite{SmorynskiC:self-reference}, never transpired back into modal logic. \emph{The point of the present paper is that combining belief updates with universal languages and self-reference leads to a curious new logical capability, whereby theories and models can be steered to assure consistency and completeness of future updates.} This capability precludes disproving current beliefs and the framework becomes \emph{belief-complete}\/ in a suitable formal sense, discussed below.

The logical framework combining belief updates and universal languages may seem unfamiliar. The main body of this paper is devoted to an attempt to describe how it arises from familiar logical frameworks. Here we try to clarify the underlying ideas. 

\paragraph{Universality.} Just like G\"odel's incompleteness theorems, our constructions of unfalsifiable beliefs are based on a \emph{universal language $\DP$}. The abstract characterization of universality, which we borrow from \cite[Ch.~2]{PavlovicD:MonCom}, is that $\DP$ comes equipped with a family of \emph{interpreters}\/ $\universal \colon \DP\times A\to B$, one for each pair of types\footnote{Each pair carries a different interpreter $\universal^{AB}$ but we elide the superscripts.} $A,B$, such that every function  $f\colon A\to B$ has a description\footnote{There may be many descriptions for each $f$ and $\enco f$ refers to an arbitrary one.} $\enco f$ in $\DP$, satisfying\footnote{The curly bracket notation allows abbreviating $\lambda a. \universal\left(p, a\right)$ to $\uev p$.} 
\bea \label{eq:univ}
f & = &  \uev{\enco f}
\eea  
This is spelled out in Sec.~\ref{Sec:moncom}. The construction in Sec.~\ref{Sec:self-fulfill} will imply that every $g\colon \DP\times A\to B$ has a fixpoint $\Gamma$, satisfying 
\bea\label{eq:fixpoint}
g(\Gamma,a) & = & \uev\Gamma (a)
\eea
Any complete programming language can be used as $\DP$. Its  interpreters support  \eqref{eq:univ} and its specializers induce \eqref{eq:fixpoint}. A sufficiently expressive software specification framework  \cite{PavlovicD:ASE01} would also fit the bill, as would a general scientific formalism \cite{Osherson:sci-inquiry}. 

\paragraph{G\"odel's incompleteness: true but unprovable statement.} G\"odel used the set of natural numbers $\NNn$ as $\LLl$, with arithmetic making it into a programming language. The concept of a programming language did not yet exist, but it came into existence through G\"odel's construction. An arithmetic expression specifying a function $f$ was encoded as a number $\enco f$ and decoded by an arithmetic function $\universal\colon\NNn\times \NNn\to \NNn$ as in \eqref{eq:univ}. A restriction of \eqref{eq:fixpoint} was proved for arithmetic predicates $p\colon \NNn\to \Bool$, where $\Bool =\{0,1\}\subset \NNn$, and a fixpoint of a predicate $g\colon \DP\times A\to \Bool$ was constructed as a predicate encoding $\enco\gamma$ satisfying\footnote{Although this discussion is semi-formal, it may be helpful to bear in mind that the equality $\{\enco\gamma\} (a) = \gamma(a)$ is \emph{extensional}: it just says that interpreting the description $\enco\gamma$ on a value $a$ always outputs the value $\gamma(a)$. But the process whereby $\{\enco\gamma\} (a)$ arrives at this value may be different from a given direct evaluation of $\gamma(a)$.}
\bea\label{eq:fixpred}
g\left(\enco\gamma,a\right) & = & \uev{\enco\gamma} (a)\ =\ \gamma(a)
\eea
To complete the incompleteness proof,  G\"odel constructed a predicate $\Prov\colon \NNn\to \Bool$ characterizing provability in formal arithmetic:
\bea\label{eq:provability}
\Prov\left(\enco p,a\right) \ \ &\iff &\ \ \vdash p(a)  
\eea
for all arithmetic predicates $p:\NNn\to \Bool$. Although proofs may be arbitrarily large, they are always finite, and if $p(a)$ has a proof, $\Prov$ will eventually find it. On the other hand, since arithmetic predicates, like all arithmetic functions, satisfy $p = \uev{\enco p}$, we also have
\bea\label{eq:provdef}
\Prov\left(\enco p,a\right) & = &  \uev{\enco p}(a) \ =\ p(a)
\eea
Setting $g(p, a) = \Prov(\enco{\neg p}, a)$  in \eqref{eq:fixpoint} induces a fixpoint $\gamma$ with
\beq
\Prov(\enco{\neg \gamma}, a) \ \ \stackrel{\eqref{eq:fixpred}}=\ \ \uev{\enco\gamma}(a)\ \ \stackrel{\eqref{eq:provdef}}= \ \ \Prov\left(\enco \gamma, a\right) 
\eeq
But \eqref{eq:provability} then implies 
\bea
\vdash \neg \gamma(a)\ \  &\iff &\ \  \vdash \gamma(a)
\eea
which means that neither $\gamma$ nor $\neg \gamma$ can be provable. On the other hand, the disjunction $\gamma \vee \neg \gamma$ is classically true. The statement $\gamma \vee \neg \gamma$ is thus true but not provable, and arithmetic is therefore incomplete. 

\paragraph{Belief completeness: universal updating.} Remarkably, the same encoding-fixpoint conundrum (\ref{eq:univ}--\ref{eq:fixpoint}), which leads to the incompleteness of static theories, also leads to the completeness of  dynamically updated theories. Updating is presented as state dependency. The function $f$ in \eqref{eq:univ} is now in the form $f:X\times A\to X\times B$ where $X$ is the state space. It may be more intuitive to think of $f$ as a \emph{process}, since it captures state changes\footnote{In automata theory, such functions are called the \emph{Mealy machines}.}. We conveniently present it as a pair $f=\left<\sta f, \out f\right>$, where $\sta f\colon X\times A\to X$ is the \emph{next state}\/ update, whereas $\out f \colon X\times A \to B$ is an $X$-indexed family of functions $\out f_{x}\colon A\to B$. The elements of the universal language $\DP$ are now construed as  \emph{belief states}. Its universality means that every observable state $x$ from any state space $X$ is expressible as a belief. The interpreters $\uuniversal\colon \DP \times A\to \DP \times B$ are also presented as pairs $\uuniversal = \left<\sta \uuniversal, \out \uuniversal\right>$, where $\sta {\uuniversal} \colon \DP\times A\to \DP$ updates the belief states whereas $\out \uuniversal \colon \DP \times A \to B$ evaluates beliefs to functions. Just like every state $x$ in $X$ determines a function $\out f_{x}\colon A\to B$, every belief $\ell$ in $\DP$ determines a function $\out{\uev\ell}\colon A\to B$, which makes predictions based on the current belief. Generalizing the fixpoint construction \eqref{eq:fixpoint}, every process $f=\left<\sta f, \out f\right>\colon X\times A\to X\times B$ now induces an assignment $\ana f\colon X\to \DP$ of beliefs to states such that
\beq \label{eq:ana}
\sta{\uev{\ana f(x)}}  =  \ana f\left(\sta f _{x}\right)\qquad \qquad \qquad \out{\uev{\ana f (x)}} = \out f_{x}
\eeq
The construction of $\ana f$ is presented in Sec.~\ref{Sec:unfalse}. Here we propose an interpretation. The second equation says that the output component of $\uuniversal$ behaves as it did in \eqref{eq:univ}: it interprets the description $\ana f(x)$ and recovers the function $\out f_{x}$ executed by the process $f$ at the state $x$. The first equation says that the interpreter $\uuniversal$ maps the $\ana f$-description of the state $x$ to the $\ana f$-description of the updated state $\sta f_{x}$:
\beq\label{eq:simul}
\prooftree
\sta f \colon x \longmapsto \sta f_{x}
\justifies
\sta\universal \colon \ana f(x) \longmapsto \ana f\left(\sta f_{x}\right)
\endprooftree
\eeq
Any state change caused by the process $f$ is thus explained by a belief update of $\ana f$ along $\uuniversal$. Interpreting the belief states $\ana f$ by the interpreter $\uuniversal$ provides belief updates that can be construed as \emph{explanations}\/ in the language $\DP$ of any state changes in the process $f$. All that can be learned about $f$ is already expressed in $\ana f$ and all state changes that may be observed will be explained by the updates anticipated by the current belief, as indicated in \eqref{eq:simul}. The belief is complete.

\paragraph{Remark.} In coalgebra and process calculus, the universal interpreters $\uuniversal \colon \DP\times A\to \DP\times B$ would be characterized as weakly final simulators \cite{PavlovicD:MonCom3}. They are universal in the sense that the same state space $\DP$ works for all types $A,B$. See \cite[Sec.~7.2]{PavlovicD:MonCom} for details and references.

\paragraph{The logic of going dynamic.} When $\DP$ is a programming language, the interpreter $\uuniversal$ interprets programs as computable functions $A\to B$, where $A$ and $B$ are types, usually predicates that allow type checking. When $\DP$ is a language of  software specifications or scientific theories construed as beliefs about the state of the world, the interpreter $\uuniversal$ updates beliefs to explain the state changes observed in explainable  processes $X\times A\to X\times B$, where $A, B$ and $X$ are state spaces. States are usually also defined by some predicates, but their purpose is not to be easy to check but to define the state changes as semantical reassignments. This is spelled out in Sec.~\ref{Sec:state}. Dynamic reassignments of meaning bring us into the realm of \emph{dynamic logic}. If the propositions from a lattice $\TTT$ are used as assertions about the states of the world or the states of our beliefs about the world, then the dynamic changes of these assertions under the influence of events from a lattice $\EEE$ can be expressed in terms of \emph{Hoare triples}
\beq\label{eq:triple} A\uev e B\eeq
saying that the event $e\in \EEE$ after the precondition $A\in \TTT$ leads to the postcondition $B\in \TTT$. The \emph{Hoare logic}\/ of such statements was developed in the late 1960s as a method for reasoning about programs. The algebra of events $\EEE$ was generated by program expressions, whereas the propositional lattice $\TTT$ was generated by formal versions of  the comments inserted by programmers into their code, to clarify the intended meanings of blocks of code \cite{FloydR:meaning,Hoare-logic}. A triple \eqref{eq:triple} would thus correspond to a block of code $e$, a comment $A$ describing the assumed state before $e$ is executed, as its precondition, and a comment $B$ describing the guaranteed state after $e$  is executed, as its postcondition. By formalizing the \emph{``assume-guarantee''}\/ reasoning of software developers, the Hoare triples provided a stepping stone into the logic of state transitions in general. The propositional algebra of dynamic logic can be viewed as a monotone map
\[ \TTT^{o}\times \EEE\times \TTT\tto{ - \uev - -} \OOO\]
where $\OOO$ is a lattice of truth values, whereas $\TTT$ and $\EEE$ are as above, and $\TTT^{o}$ is $\TTT$ with the opposite order. If the lattice $\TTT$ is complete, then each event $e\in \EEE$ induces a Galois connection
\[ A\rtimes e \vdash B \ \ \iff\ \ A\uev e B\ \ \iff\ \ A\vdash [e]B\]
determining a \emph{dynamic modality}\/ $[e]\colon\TTT\to\TTT$ for every $e\in \EEE$ \cite{PrattV:dynamic}. The induced interior operation $\left([e]B\right)\rtimes E\vdash B$ says that $[e]B$ is the weakest precondition that guarantees $B$ after $e$. The induced closure $A\vdash [e]\left(A\rtimes e\right)$ says that $A\rtimes e$ is the strongest postcondition that can be guaranteed by the assumption $A$ before $e$. In addition to formal program annotations, dynamic logic found many other uses and interpretations \cite{BenthemJ:lambdas,Ditmarsch:DEL,Kozen:dynamic-book}. Here we use it as a backdrop for the coevolution of theories and their interpretations.

\paragraph{Updating completeness.} In static logic, a theory is complete when all statements true in a reference model are provable in the theory. In dynamic logic, the model changes dynamically and the true statements vary. There are different ways in which the notion of completeness can be generalized for dynamic situations. The notion of completeness that seems to be of greatest practical interest is the requirement that the theory and the model can be dynamically adapted to each other: the theory can be updated to make provable some true statements or the model can be updated to make true some false statements.   This requirement covers both the theory updates in science and the model updates by self-fulfilling  and belief-building announcements  in various non-sciences. The logical frameworks satisfying such completeness requirements allow for matching current beliefs and future states.

\section{World as a monoidal category}
\label{Sec:world}
% !TEX root = 00-wollic.tex

\subsection{State spaces as objects}\label{Sec:state}
In computation, a state is a family of typed variables with a partial assignment of values. In science, a state is a family of observables, some with expected values. Formally, a state can be viewed as a family of predicates, or a theory in first-order logic,  with a specified model. Both can be presented in the standard Tarskian format, where a theory is a quadruple of sorts, operations, predicates, and axioms, and its  interpretation is an inductively defined model  \cite{Chang-Keisler}. 

\paragraph{Theories as sketches.} In this extended abstract, theories are  presented as categorical \emph{sketches}\/ and their models are specified in extended functorial semantics \cite{AdamekJ:locpac,Ehresmann-Bastiani:sketches,LairC:modelables,LairC:esquissables,MakkaiM:acccfc}. While this may not be the most popular view, it is succinct enough to fit into the available space. The main constructions, presented in Sec.~\ref{Sec:moncom}--\ref{Sec:unfalse}, do not depend on the choice of presentation. The reader could thus skip to Sec.~\ref{Sec:string} and come back as needed. 

\begin{definition}\label{def:state}
A \emph{clone} $\Sigma$ is a cartesian category\footnote{We stick with the traditional terminology where a category is cartesian when it has cartesian products. The cartesian product preserving functors are abbreviated to \emph{cartesian functors}. This clashes with the standard terminology for morphisms between fibrations, but fibrations do not come about in this paper.} freely generated by sorts, operations, and equational axioms of a logical theory. A \emph{theory}\/ is a pair $\Theta=<\Sigma,\Gamma>$, where $\Sigma$ is a clone and $\Gamma$ is a set of cones and cocones in $\Sigma$, capturing the general axioms\footnote{Equational axioms could be subsumed among cones and cocones, and omitted from $\Sigma$, which would boil down to the free category generated by sorts and operations.} of the logical theory. A \emph{model}\/ of $\Theta$ is a cartesian functor $\MMM\colon \Sigma \to \Set$ mapping the $\Gamma$-cones into limit cones and the $\Gamma$-cocones into colimit cocones. A \emph{state of belief} (or \emph{belief state}) is a triple 
\bear
A & = & \left<\Sigma_{A},\Gamma_{A}, \MMM_{A}\right> 
\eear
where $\Theta_{A} = \left<\Sigma_{A},\Gamma_{A}\right>$ is a theory and $\MMM_{A}$ its model in a category $\Set$ of sets and functions. An\/ element of the model $\MMM_{A}$ is called an \emph{observable} of the state $A$.
\end{definition}

\paragraph{States of $A$ as extensions of $\MMM_{A}$.} The reference model $\MMM_A$ determines the notion of truth in the state space $A$. It expresses properties that may not be proved in the theory $\Theta_{A}$ or even effectively specified\footnote{E.g., the set of all true statements of Peano arithmetic is expressed by its standard model, but most of them cannot be described effectively.}. The reference model $\MMM_{A}$ should thus not be thought of as a single object of the category of all models of $\Theta_{A}$ but as the (accessible) subcategory of model extensions of $\MMM_A$. These model extensions are the states of the state space $A$. The structure of a state space can be further refined to capture other features of  theories in science and engineering, including their statistical and complexity-theoretic valuations \cite{RissanenJ:MDL,WallaceCS:MML}. While such refinements have no direct impact on our considerations, they signal that we are in the realm of \emph{inductive}\/ inference, which may feel unusual for the Tarskian framework of static logic, normally concerned with deductive aspects. The fact that the theory $\Theta_{A}$ has a model $\MMM_{A}$ implies that it is logically consistent but it does not imply that it is true within an external frame of reference, a \emph{``reality''}\/ that may drive the state transitions, i.e. the processes of extending and reinterpreting theories. The intuition is that the states in the space $A$ are observables that may never be observed, since $\MMM_{A}$ may be incompatible with the actual observations. The theory $\Theta_{A}$ may be consistent but wrong.

\paragraph{Examples} of state spaces include logical theories with standard models that arise not only in natural sciences but also in social systems, as policy formalizations.  A software specification with a reference implementation can also be viewed as a state space. Updates and evolution of a software system can then be analyzed using a higher-order dynamic logic \cite{PavlovicD:JAMP}. The functorial semantics view was spelled out in \cite{PavlovicD:FOPS}, used in a software synthesis tool \cite{PavlovicD:AMAST08,PavlovicD:ASE01,PavlovicD:SDR}, and applied in algorithm design \cite{PavlovicD:ManaFest,PavlovicD:MPC10}. 

\subsection{Transitions as morphisms}\label{Sec:transition}
Intuitively, a transition $f$ from a state space $A$ to a state space $B$ is a specification that induces a transition from any $A$-state to a $B$-state. We first consider the transitions arising from reinterpreting theories and then expand to modifying the reference models.

\begin{definition}\label{Def:interpretable}
An \emph{interpretation}\/ of state space $A$ in a state space $B$ is a logical interpretation of the theory $\Theta_{B}=<\Sigma_{B},\Gamma_{B}>$ in the theory $\Theta_{A} =<\Sigma_{A},\Gamma_{A}>$ which reduces the reference model $\MMM_{A}$ to $\MMM_{B}$. More precisely, an interpretation  $f\colon A\to B$ is a cartesian functor $f\colon \Sigma_{A}\ot %\ooot{\Gamma_{A}}{\Gamma_{B}} 
\Sigma_{B}$ mapping $\Gamma_{B}$-(co)cones to $\Gamma_{A}$-(co)cones according to a given assignment $f_\Gamma\colon \Gamma_{A}\ot \Gamma_{B}$ and making the following diagram commute
\beq\label{eq:def-intepretable}
\begin{tikzar}{}
\Sigma_{A}\ar{dr}[description]{\MMM_{A}} \&\& \Sigma_{B}\ar{dl}[description]{\MMM_{B}} \ar[bend right]{ll}[swap]{f}\\
\& \Set
\end{tikzar}
\eeq
The models $\MMM_A$ and $\MMM_B$ map the (co)cones from $\Gamma_A$ and $\Gamma_B$ to (co)limits of sets, as required by Def.~\ref{def:state}. \end{definition}

\paragraph{Interpretations as assignments.} The structure of interpretations of software specifications and the method to compose them were spelled out in  \cite{PavlovicD:FOPS,PavlovicD:ASE01}. Since software specifications are finite, an interpretation $f\colon \Sigma_{A}\ot \Sigma_{B}$ boils down to a tuple of assignments 
\[x_{1}:=t_{1}\ ;\  x_{2}:=t_{2}\ ;\ldots;\  x_{n}:=t_{n}\]
of terms $\vec t=<t_{1}, t_{2},\ldots, t_{n}>$ from $\Sigma_{A}$ to variables $\vec x=<x_{1}, x_{2},\ldots, x_{n}>$ from $ \Sigma_{B}$ in such a way that, for each axiom $\gamma\in \Gamma_{B}$, the substitution instance 
\bear
f(\gamma) &= & [\vec{x}:=\vec {t}]\gamma
\eear  
is a theorem derivable from the axioms in $\Gamma_{A}$. In Hoare logic \cite{Hoare-logic}, a state transition $f\colon \Sigma_{A}\ot \Sigma_{B}$ is presented as a triple $\Theta_{A}\uev{\vec{x}:=\vec {t}} \Theta_{B}$.  By definition, this triple is valid if and only if $\Theta_{A}\vdash [\vec{x}:=\vec {t}]\Theta_{B}$, where $[\vec{x}:=\vec {t}]\Theta_{B}$ is the result substituting the $\Theta_{A}$-terms $\vec t$ for $\Theta_{B}$-variables $\vec x$ in all axioms $\gamma\in \Gamma_{B}$. Condition \eqref{eq:def-intepretable} moreover requires that this theory interpretation recovers the model $\MMM_{B}$ from the model $\MMM_{A}$. 

In general, however, it is not always possible to transform all computational states annotated at all relevant program points into one another by mere substitutions. That is why Hoare logic does not boil down to the assignment clause, but specifies the meaning of other program constants in other clauses, which can be viewed as more general state transitions.

\begin{definition}\label{Def:explainable}
A \emph{state transition}\/  $f\colon A\to B$ is a cartesian functor  $f\colon \Theta_{A}\ot %\ooot{\Gamma_{A}}{\Gamma_{B}} 
\Theta_{B}$ mapping $\Gamma_{B}$-(co)cones to $\Gamma_{A}$-(co)cones according to a given assignment $f_\Gamma\colon \Gamma_{A}\ot \Gamma_{B}$ and moreover making the following diagram commute
\beq\label{eq:def-explainable}
\begin{tikzar}{}
\Theta_{A}\ar{dr}[description]{\overline\MMM_{A}} \&\& \Theta_{B}\ar{dl}[description]{\overline\MMM_{B}} \ar[bend right]{ll}[swap]{f}\\
\& \Set
\end{tikzar}
\eeq
where $\overline\MMM_A$ is the extension of $\MMM_A$ along the completion $\Sigma_A\inclusion \Theta_A$ of $\Sigma_A$ under the limits and colimits generated by $\Gamma_A$; ditto for $\overline\MMM_B$.
\end{definition}

\paragraph{General sketches.} In Def.~\ref{Def:interpretable}, theories were presented as pairs $\Theta = <\Sigma,\Gamma>$, where the category $\Sigma$ is comprised of sorts, operations, and equations of the theory, whereas the cones and the cocones in $\Gamma$ specify its predicates and axioms. In Def.~\ref{Def:explainable}, a theory $\Theta$ is presented as the category obtained by completing $\Sigma$ under the limits and the colimits specified by $\Gamma$. This general sketch, with the family of limit cones and colimit cocones from $\Gamma$, is now denoted $\Theta$, by abuse of notation. A detailed construction of this sketch can be found in \cite[\S4.2--3]{MakkaiM:acccfc}. It is a canonical view of the theory derived in the signature $\Sigma$ from the axioms $\Gamma$. Since the category $\Theta$ is the $\Gamma$-completion of $\Sigma$, any functor $\MMM:\Sigma\to\Set$ mapping the $\Gamma$-(co)cones in $\Sigma$ to (co)limit (co)cones in $\Set$ has a unique $\Gamma$-preserving extension $\overline \MMM\colon \Theta\to \Set$. These extensions are displayed in \eqref{eq:def-intepretable}. The upshot of saturating the sketches from Def.~\ref{Def:interpretable} in the form $\Theta=<\Sigma,\Gamma>$ to the general sketches over $\Theta$ in Def.~\ref{Def:explainable} is that the general explainable transitions are now simply the structure-preserving functors displayed in \eqref{eq:def-intepretable}.

\subsection{Monoidal category of state spaces and transitions}
Let 
\begin{itemize}
\item $\UUU$ be the category of state spaces from Def.~\ref{def:state} and transitions from Def.~\ref{Def:explainable}, and let
\item $\tot \UUU$ be the category of state spaces from Def.~\ref{def:state} and interpretations from Def.~\ref{Def:interpretable}.
\end{itemize}
In both cases, the monoidal structure is induced by the disjoint unions of theories:
\bea
A\otimes B & = & \Big<\Sigma_{A}+\Sigma_{B}\, ,\,  \Gamma_{A}+\Gamma_{B}\, ,\,  [\MMM_{A}+\MMM_{B}]
\Big>
\eea
where $\MMM_{A\otimes B} = [\MMM_{A}+\MMM_{B}]\colon \Sigma_{A}+\Sigma_{B} \tto{\Gamma_{A\otimes B}}\Set$ maps $\Sigma_{A}$ like $\MMM_{A}$ and $\Sigma_{B}$ like $\MMM_{B}$. The tensor unit is $I = \left<\bot, \bot, \emptyset\right>$, where the truth value $\bot$ denotes the inconsistent theory or sketch, its only axiom, and $\emptyset$ is its empty model. It obviously satisfies $I\otimes A = A = A\otimes I$. The associativity of the tensor $\otimes$ follows from the associativity of the disjoint union $+$. The arrow part of $\otimes$ is induced by the disjoint unions as coproducts. The coproduct structure equips every state space $A$ with a cartesian comonoid structure
\begin{gather}\label{eq:dataserv-text}
 \ \ A\otimes A\  \oot{\ \ \ \ \Delta\ \  \ \ } A \tto{\ \  \ \scun\ \ \ } I\\
\Sigma_{A}+\Sigma_{A}\tto{\ [\id,\id]\ } \Sigma_{A}\oot{\ \ \bot\ \ } \bot\notag
\end{gather}
This provides a categorical mechanism for cloning and erasing states, which makes some observations repeatable and deletable, as required for testing in science and software engineering. However, $\UUU$ is not a cartesian category, and $\otimes$ is not a cartesian product, because some transitions $f:A\to B$ do not in general boil down to functors $\Sigma_{A}\ot \Sigma_{B}$, but only to functors $\Theta_{A}\ot \Sigma_{B}$, where $\Theta_A$ is a completion of $\Sigma_A$ under the $\Gamma_A$-(co)-limits. Intuitively, this means that the axioms of the theory $\Theta_{B}$ may not be interpreted as axioms of $\Theta_{A}$, but may be mapped into theorems, which only arise in the $\Gamma_{A}$-completion. This captures the uncloneable and undeletable states that arise in many sciences, including physics of very small or very large (quantum or cosmological) and economics. The only transitions that preserve the cartesian structure \eqref{eq:dataserv-text} are the interpretations $f:A\to B$, with the underlying functors $\Sigma_{A}\ot \Sigma_{B}$. They form the category $\tot\UUU$, which is the largest cartesian subcategory of $\UUU$. If the states $\alpha \in \UUU(I,A)$ are thought of as observables, the states $a\in \tot \UUU(I,A)$ are the actual observations.

\section{String diagrams}
\label{Sec:string}
% !TEX root = 00-wollic.tex

Constructions in monoidal categories yield to insightful presentations in terms of string diagrams \cite[Ch.~1]{Joyal-Street:geometry,PavlovicD:MonCom}. We will need them to present the constructions like \eqref{eq:fixpoint} and in particular  \eqref{eq:ana}. While commutative diagrams like \eqref{eq:def-intepretable} display compositions of morphisms and abbreviate their equations, string diagrams display \emph{de}\/compositions of morphisms. Monoidal categories come with two composition operations: the categorical (sequential) morphism composition $\circ$ and the monoidal (parallel) composition $\otimes$. The former is drawn along the vertical axis, the latter along the horizontal axis. The objects are drawn as strings, the morphisms as boxes. A morphism $A\tto f B$ is presented as a box $f$ with a string $A$ hanging from the bottom and a string $B$ sticking out from the top. The identities are presented as invisible boxes: the identity on $A$ is just the string $A$. The unit type $I$ is presented as the invisible string. There are thus boxes with no strings attached. The composite morphism $g\circ f =(A\tto f B\tto g C)$ is drawn bottom-up, by hanging the box $f$ on the string $B$ under the box $g$. The monoidal composition is presented as the horizontal adjacency: the composite $(g\circ f)\otimes (s\circ t)$ is drawn by placing the boxes $g\circ f$ next to the boxes for $s\circ t$:

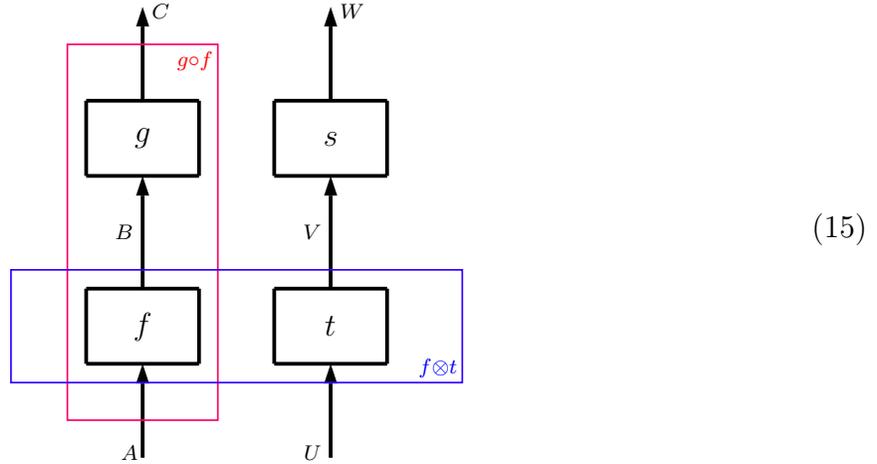
\beq\label{eq:godement}
\begin{split}
\newcommand{\machine}{$f$}
\newcommand{\gee}{$g$}
\newcommand{\kee}{$s$}
\newcommand{\hee}{t}
\newcommand{\nameslang}{\scriptstyle B}
\newcommand{\seqcompp}{\scriptstyle {\color{red}g\circ f}}
\newcommand{\parcompp}{\scriptstyle {\color{blue}f\otimes t}}
\newcommand{\inputt}{\scriptstyle A} 
\newcommand{\outpt}{$\scriptstyle C$}
\newcommand{\otherinputt}{\scriptstyle U}
\newcommand{\otheroutpt}{\scriptstyle V} 
\newcommand{\outpttt}{$\scriptstyle W$}
\def\JPicScale{.5}
\ifx\JPicScale\undefined\def\JPicScale{1}\fi
\psset{unit=\JPicScale mm}
\psset{linewidth=0.3,dotsep=1,hatchwidth=0.3,hatchsep=1.5,shadowsize=1,dimen=middle}
\psset{dotsize=0.7 2.5,dotscale=1 1,fillcolor=black}
\psset{arrowsize=1 2,arrowlength=1,arrowinset=0.25,tbarsize=0.7 5,bracketlength=0.15,rbracketlength=0.15}
\begin{pspicture}(0,0)(115,115)
\psline[linewidth=1](15,40)(45,40)
\psline[linewidth=1](15,40)(15,20)
\psline[linewidth=1](45,20)(45,40)
\psline[linewidth=1,arrowsize=1.5 2,arrowlength=1.5,arrowinset=0]{<-}(30,70)(30,40)
\rput[r](27.5,55){$\nameslang$}
\rput[r](28.75,-3.75){$\inputt$}
\psline[linewidth=1](15,20)(45,20)
\psline[linewidth=1,arrowsize=1.5 2,arrowlength=1.5,arrowinset=0]{<-}(30,20)(30,-5)
\rput(30,30){\machine}
\psline[linewidth=1](15,90)(45,90)
\psline[linewidth=1](15,90)(15,70)
\psline[linewidth=1](45,70)(45,90)
\psline[linewidth=1,arrowsize=1.5 2,arrowlength=1.5,arrowinset=0]{<-}(30,115)(30,90)
\psline[linewidth=1](15,70)(45,70)
\rput(30,80){\gee}
\rput[l](32.5,113.75){\outpt}
\psline[linewidth=1](65,40)(95,40)
\psline[linewidth=1](65,40)(65,20)
\psline[linewidth=1](95,20)(95,40)
\psline[linewidth=1,arrowsize=1.5 2,arrowlength=1.5,arrowinset=0]{<-}(80,70)(80,40)
\psline[linewidth=1](65,20)(95,20)
\psline[linewidth=1,arrowsize=1.5 2,arrowlength=1.5,arrowinset=0]{<-}(80,20)(80,-5)
\rput[r](77.5,-3.75){$\otherinputt$}
\rput[r](77.5,55){$\otheroutpt$}
\rput(80,30){$\hee$}
\newrgbcolor{userLineColour}{1 0 0.4}
\pspolygon[linewidth=0.45,linecolor=userLineColour](10,105)(50,105)(50,5)(10,5)
\newrgbcolor{userLineColour}{0.2 0 1}
\pspolygon[linewidth=0.45,linecolor=userLineColour](-5,45)(115,45)(115,15)(-5,15)
\rput[br](113.75,16.25){$\parcompp$}
\rput[tr](48.75,103.12){$\seqcompp$}
\psline[linewidth=1](65,90)(65,70)
\psline[linewidth=1](95,70)(95,90)
\psline[linewidth=1,arrowsize=1.5 2,arrowlength=1.5,arrowinset=0]{<-}(80,115)(80,90)
\psline[linewidth=1](65,70)(95,70)
\psline[linewidth=1](65,90)(95,90)
\rput(80,80){\kee}
\rput[l](82.5,113.75){\outpttt}
\end{pspicture}

\end{split}
\eeq

The middle-two-interchange law $(g\circ f)\otimes(s\circ t) =  (g\otimes s)\circ(f\otimes t)$ corresponds to the two ways of reading the diagram: vertical-first and horizontal-first, marked by the red and the blue rectangle respectively. The string diagrams corresponding to the cartesian comonoids \eqref{eq:dataserv-text} are 
\beq\label{eq:dataserv}
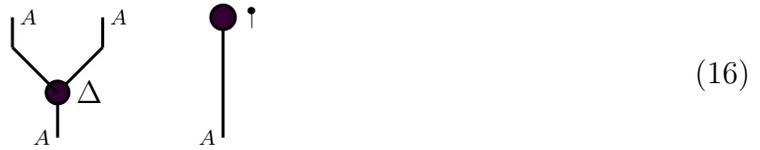
\begin{split}
\newcommand{\AAh}{\scriptstyle A}
\newcommand{\ccopy}{\cmn}
\newcommand{\delete}{\scun}
\def\JPicScale{.4} 
\ifx\JPicScale\undefined\def\JPicScale{1}\fi
\psset{unit=\JPicScale mm}
\psset{linewidth=0.3,dotsep=1,hatchwidth=0.3,hatchsep=1.5,shadowsize=1,dimen=middle}
\psset{dotsize=0.7 2.5,dotscale=1 1,fillcolor=black}
\psset{arrowsize=1 2,arrowlength=1,arrowinset=0.25,tbarsize=0.7 5,bracketlength=0.15,rbracketlength=0.15}
\begin{pspicture}(0,0)(77.5,44.07)
\psline[linewidth=1](70,40)(70,0)
\newrgbcolor{userFillColour}{0.2 0 0.2}
\rput{0}(70,40){\psellipse[linewidth=1,fillcolor=userFillColour,fillstyle=solid](0,0)(4.07,-4.07)}
\psline[linewidth=1](30,30)(15,15)
\psline[linewidth=1](15,15)(15,0)
\newrgbcolor{userFillColour}{0.2 0 0.2}
\rput{0}(15,15){\psellipse[linewidth=1,fillcolor=userFillColour,fillstyle=solid](0,0)(3.79,-3.79)}
\psline[linewidth=1](0,30)(15,15)
\rput[r](12.5,0){$\AAh$}
\rput[r](67.5,0){$\AAh$}
\psline[linewidth=1](0,40)(0,30)
\psline[linewidth=1](30,40)(30,30)
\rput[l](32.5,40){$\AAh$}
\rput[l](2.5,40){$\AAh$}
\rput[l](77.5,40){$\delete$}
\rput[l](21.25,15){$\ccopy$}
\end{pspicture}

\end{split}
\eeq
The equations that make them into commutative comonoids look like this:
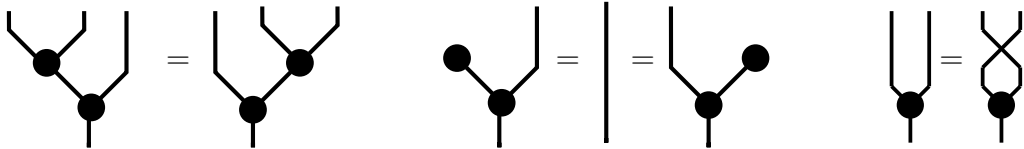
\begin{alignat}{11}
\def\JPicScale{1} %%Created by jPicEdt 1.4.1_03: mixed JPIC-XML/LaTeX format
\ifx\JPicScale\undefined\def\JPicScale{1}\fi
\psset{unit=\JPicScale mm}
\psset{linewidth=0.3,dotsep=1,hatchwidth=0.3,hatchsep=1.5,shadowsize=1,dimen=middle}
\psset{dotsize=0.7 2.5,dotscale=1 1,fillcolor=black}
\psset{arrowsize=1 2,arrowlength=1,arrowinset=0.25,tbarsize=0.7 5,bracketlength=0.15,rbracketlength=0.15}
\begin{pspicture}(0,0)(15,7.5)
\psline[linewidth=0.55](5.62,-0.62)
(10.62,-5.62)
(9.38,-5.62)(10,-5.62)
\rput{0}(4.38,0.62){\psellipse[linewidth=0.55,fillstyle=solid](0,0)(1.56,-1.57)}
\rput{0}(10.32,-5.31){\psellipse[linewidth=0.55,fillstyle=solid](0,0)(1.56,-1.57)}
\psline[linewidth=0.55](15,7.5)
(15,5)
(15,-0.63)(11.25,-4.38)
\psline[linewidth=0.55](9.37,7.5)
(9.37,6.87)
(9.38,5)(5.63,1.25)
\psline[linewidth=0.55](-0.63,7.5)
(-0.63,6.87)
(-0.62,5)(3.13,1.25)
\psline[linewidth=0.55](10,-6.88)
(10,-10)
(10,-10.62)(10,-10)
\end{pspicture}
\   &\ \ =\ \   \def\JPicScale{1} %%Created by jPicEdt 1.4.1_03: mixed JPIC-XML/LaTeX format
\ifx\JPicScale\undefined\def\JPicScale{1}\fi
\psset{unit=\JPicScale mm}
\psset{linewidth=0.3,dotsep=1,hatchwidth=0.3,hatchsep=1.5,shadowsize=1,dimen=middle}
\psset{dotsize=0.7 2.5,dotscale=1 1,fillcolor=black}
\psset{arrowsize=1 2,arrowlength=1,arrowinset=0.25,tbarsize=0.7 5,bracketlength=0.15,rbracketlength=0.15}
\begin{pspicture}(0,0)(15.63,8.12)
\rput{0}(10.62,0.62){\psellipse[linewidth=0.55,fillstyle=solid](0,0)(1.57,-1.56)}
\psline[linewidth=0.55](15.62,8.12)
(15.62,7.5)
(15.63,5.62)(11.88,1.87)
\psline[linewidth=0.55](5.62,8.12)
(5.62,7.5)
(5.63,5.62)(9.38,1.87)
\rput{0}(4.38,-5.62){\psellipse[linewidth=0.55,fillstyle=solid](0,0)(1.57,-1.57)}
\psline[linewidth=0.55](-0.63,7.5)
(-0.62,1.25)
(-0.62,-0.63)(3.13,-4.38)
\pscustom[linewidth=0.55]{\psline(9.38,-0.62)(4.38,-5.62)
\psbezier(4.38,-5.62)(4.38,-5.62)(4.38,-5.62)
\psbezier(4.38,-5.62)(4.38,-5.62)(4.38,-5.62)
}
\psline[linewidth=0.55](4.38,-6.88)
(4.38,-10)
(4.38,-10.62)(4.38,-10)
\end{pspicture}
 &&\qquad\qquad&&& 
\def\JPicScale{1} 
%%Created by jPicEdt 1.4.1_03: mixed JPIC-XML/LaTeX format
%%Sat May 15 12:55:58 GMT-10:00 2021
%%Begin JPIC-XML
%<?xml version="1.0" standalone="yes"?>
%<jpic x-min="-2.19" x-max="10" y-min="-10.62" y-max="8.12" auto-bounding="true">
%<multicurve fill-style= "none"
%	 stroke-width= "0.55"
%	 points= "(0.62,0);(0.62,0);(5.62,-5);(5.62,-5);(5.62,-5);(4.38,-5);
%	(4.38,-5);(4.38,-5);(5,-5);(5,-5)"
%	 />
%<ellipse p3= "(-2.19,2.81)"
%	 p2= "(-2.19,-0.31)"
%	 fill-style= "solid"
%	 p1= "(0.94,-0.31)"
%	 stroke-width= "0.55"
%	 closure= "open"
%	 angle-end= "0"
%	 angle-start= "0"
%	 />
%<ellipse p3= "(3.75,-3.13)"
%	 p2= "(3.75,-6.25)"
%	 fill-style= "solid"
%	 p1= "(6.88,-6.25)"
%	 stroke-width= "0.55"
%	 closure= "open"
%	 angle-end= "0"
%	 angle-start= "0"
%	 />
%<multicurve fill-style= "none"
%	 stroke-width= "0.55"
%	 points= "(10,8.12);(10,8.12);(10,5.62);(10,5.62);(10,5.62);(10,0);
%	(10,0);(10,0);(6.25,-3.75);(6.25,-3.75)"
%	 />
%<multicurve fill-style= "none"
%	 stroke-width= "0.55"
%	 points= "(5,-6.25);(5,-6.25);(5,-10);(5,-10);(5,-10);(5,-10.62);
%	(5,-10.62);(5,-10.62);(5,-10);(5,-10)"
%	 />
%</jpic>
%%End JPIC-XML
%PSTricks content-type (pstricks.sty package needed)
%Add \usepackage{pstricks} in the preamble of your LaTeX file
%You can rescale the whole picture (to 80% for instance) by using the command \def\JPicScale{0.8}
\ifx\JPicScale\undefined\def\JPicScale{1}\fi
\psset{unit=\JPicScale mm}
\psset{linewidth=0.3,dotsep=1,hatchwidth=0.3,hatchsep=1.5,shadowsize=1,dimen=middle}
\psset{dotsize=0.7 2.5,dotscale=1 1,fillcolor=black}
\psset{arrowsize=1 2,arrowlength=1,arrowinset=0.25,tbarsize=0.7 5,bracketlength=0.15,rbracketlength=0.15}
\begin{pspicture}(0,0)(10,8.12)
\psline[linewidth=0.55](0.62,0)
(5.62,-5)
(4.38,-5)(5,-5)
\rput{0}(-0.62,1.25){\psellipse[linewidth=0.55,fillstyle=solid](0,0)(1.57,-1.56)}
\rput{0}(5.31,-4.69){\psellipse[linewidth=0.55,fillstyle=solid](0,0)(1.57,-1.56)}
\psline[linewidth=0.55](10,8.12)
(10,5.62)
(10,0)(6.25,-3.75)
\psline[linewidth=0.55](5,-6.25)
(5,-10)
(5,-10.62)(5,-10)
\end{pspicture}
  & \ = \ &\ \, \def\JPicScale{1} %%Created by jPicEdt 1.4.1_03: mixed JPIC-XML/LaTeX format
%%Sat May 15 12:55:13 GMT-10:00 2021
%%Begin JPIC-XML
%<?xml version="1.0" standalone="yes"?>
%<jpic x-min="0" x-max="0" y-min="-10" y-max="8.74" auto-bounding="true">
%<multicurve fill-style= "none"
%	 stroke-width= "0.55"
%	 points= "(0,8.74);(0,8.74);(0,-9.38);(0,-9.38);(0,-9.38);(0,-10);
%	(0,-10);(0,-10);(0,-9.38);(0,-9.38)"
%	 />
%</jpic>
%%End JPIC-XML
%PSTricks content-type (pstricks.sty package needed)
%Add \usepackage{pstricks} in the preamble of your LaTeX file
%You can rescale the whole picture (to 80% for instance) by using the command \def\JPicScale{0.8}
\ifx\JPicScale\undefined\def\JPicScale{1}\fi
\psset{unit=\JPicScale mm}
\psset{linewidth=0.3,dotsep=1,hatchwidth=0.3,hatchsep=1.5,shadowsize=1,dimen=middle}
\psset{dotsize=0.7 2.5,dotscale=1 1,fillcolor=black}
\psset{arrowsize=1 2,arrowlength=1,arrowinset=0.25,tbarsize=0.7 5,bracketlength=0.15,rbracketlength=0.15}
\begin{pspicture}(0,0)(0,8.74)
\psline[linewidth=0.55](0,8.74)
(0,-9.38)
(0,-10)(0,-9.38)
\end{pspicture}
 &\ \, = \ \ &  \def\JPicScale{1} %%Created by jPicEdt 1.4.1_03: mixed JPIC-XML/LaTeX format
%%Sat May 15 12:56:33 GMT-10:00 2021
%%Begin JPIC-XML
%<?xml version="1.0" standalone="yes"?>
%<jpic x-min="-0.63" x-max="12.19" y-min="-10.62" y-max="8.12" auto-bounding="true">
%<ellipse p3= "(9.06,2.81)"
%	 p2= "(9.06,-0.31)"
%	 fill-style= "solid"
%	 p1= "(12.19,-0.31)"
%	 stroke-width= "0.55"
%	 closure= "open"
%	 angle-end= "0"
%	 angle-start= "0"
%	 />
%<ellipse p3= "(2.81,-3.44)"
%	 p2= "(2.81,-6.56)"
%	 fill-style= "solid"
%	 p1= "(5.94,-6.56)"
%	 stroke-width= "0.55"
%	 closure= "open"
%	 angle-end= "0"
%	 angle-start= "0"
%	 />
%<multicurve fill-style= "none"
%	 stroke-width= "0.55"
%	 points= "(4.38,-6.25);(4.38,-6.25);(4.38,-10);(4.38,-10);(4.38,-10);(4.38,-10.62);
%	(4.38,-10.62);(4.38,-10.62);(4.38,-10);(4.38,-10)"
%	 />
%<multicurve fill-style= "none"
%	 stroke-width= "0.55"
%	 points= "(-0.63,8.12);(-0.63,8.12);(-0.62,1.87);(-0.62,1.87);(-0.62,1.87);(-0.62,0);
%	(-0.62,0);(-0.62,0);(3.13,-3.75);(3.13,-3.75)"
%	 />
%<multicurve fill-style= "none"
%	 stroke-width= "0.55"
%	 points= "(9.38,0);(9.38,0);(4.38,-5);(4.38,-5);(4.38,-5);(4.38,-5);
%	(4.38,-5);(4.38,-5);(4.38,-5);(4.38,-5)"
%	 />
%</jpic>
%%End JPIC-XML
%PSTricks content-type (pstricks.sty package needed)
%Add \usepackage{pstricks} in the preamble of your LaTeX file
%You can rescale the whole picture (to 80% for instance) by using the command \def\JPicScale{0.8}
\ifx\JPicScale\undefined\def\JPicScale{1}\fi
\psset{unit=\JPicScale mm}
\psset{linewidth=0.3,dotsep=1,hatchwidth=0.3,hatchsep=1.5,shadowsize=1,dimen=middle}
\psset{dotsize=0.7 2.5,dotscale=1 1,fillcolor=black}
\psset{arrowsize=1 2,arrowlength=1,arrowinset=0.25,tbarsize=0.7 5,bracketlength=0.15,rbracketlength=0.15}
\begin{pspicture}(0,0)(12.19,8.12)
\rput{0}(10.62,1.25){\psellipse[linewidth=0.55,fillstyle=solid](0,0)(1.56,-1.56)}
\rput{0}(4.38,-5){\psellipse[linewidth=0.55,fillstyle=solid](0,0)(1.57,-1.56)}
\psline[linewidth=0.55](4.38,-6.25)
(4.38,-10)
(4.38,-10.62)(4.38,-10)
\psline[linewidth=0.55](-0.63,8.12)
(-0.62,1.87)
(-0.62,0)(3.13,-3.75)
\pscustom[linewidth=0.55]{\psline(9.38,0)(4.38,-5)
\psbezier(4.38,-5)(4.38,-5)(4.38,-5)
\psbezier(4.38,-5)(4.38,-5)(4.38,-5)
}
\end{pspicture}
&&\qquad\qquad&&&
\def\JPicScale{1} %%Created by jPicEdt 1.4.1_03: mixed JPIC-XML/LaTeX format
%%Thu Mar 16 14:52:17 GMT-10:00 2023
%%Begin JPIC-XML
%<?xml version="1.0" standalone="yes"?>
%<jpic x-min="0" x-max="5" y-min="-10" y-max="7.5" auto-bounding="true">
%<ellipse fill-style= "solid"
%	 stroke-width= "0.5"
%	 p3= "(0.93,-3.44)"
%	 p2= "(0.93,-6.56)"
%	 p1= "(4.06,-6.56)"
%	 closure= "open"
%	 angle-end= "0"
%	 angle-start= "0"
%	 />
%<multicurve fill-style= "none"
%	 stroke-width= "0.5"
%	 points= "(2.5,-5);(2.5,-5);(2.5,-10);(2.5,-10)"
%	 />
%<multicurve fill-style= "none"
%	 stroke-width= "0.5"
%	 points= "(5,-2.5);(5,-2.5);(2.5,-5);(2.5,-5)"
%	 />
%<multicurve fill-style= "none"
%	 stroke-width= "0.5"
%	 points= "(0,-2.5);(0,-2.5);(2.5,-5);(2.5,-5)"
%	 />
%<multicurve fill-style= "none"
%	 stroke-width= "0.5"
%	 points= "(5,7.5);(5,7.5);(5,-2.5);(5,-2.5)"
%	 />
%<multicurve fill-style= "none"
%	 stroke-width= "0.5"
%	 points= "(0,7.5);(0,7.5);(0,-2.5);(0,-2.5)"
%	 />
%</jpic>
%%End JPIC-XML
%PSTricks content-type (pstricks.sty package needed)
%Add \usepackage{pstricks} in the preamble of your LaTeX file
%You can rescale the whole picture (to 80% for instance) by using the command \def\JPicScale{0.8}
\ifx\JPicScale\undefined\def\JPicScale{1}\fi
\psset{unit=\JPicScale mm}
\psset{linewidth=0.3,dotsep=1,hatchwidth=0.3,hatchsep=1.5,shadowsize=1,dimen=middle}
\psset{dotsize=0.7 2.5,dotscale=1 1,fillcolor=black}
\psset{arrowsize=1 2,arrowlength=1,arrowinset=0.25,tbarsize=0.7 5,bracketlength=0.15,rbracketlength=0.15}
\begin{pspicture}(0,0)(5,7.5)
\rput{0}(2.49,-5){\psellipse[linewidth=0.5,fillstyle=solid](0,0)(1.56,-1.56)}
\psline[linewidth=0.5](2.5,-5)(2.5,-10)
\psline[linewidth=0.5](5,-2.5)(2.5,-5)
\psline[linewidth=0.5](0,-2.5)(2.5,-5)
\psline[linewidth=0.5](5,7.5)(5,-2.5)
\psline[linewidth=0.5](0,7.5)(0,-2.5)
\end{pspicture}
 &\   = \   \def\JPicScale{1} %%Created by jPicEdt 1.4.1_03: mixed JPIC-XML/LaTeX format
%%Thu Mar 16 14:52:39 GMT-10:00 2023
%%Begin JPIC-XML
%<?xml version="1.0" standalone="yes"?>
%<jpic x-min="0" x-max="5" y-min="-10" y-max="7.5" auto-bounding="true">
%<ellipse fill-style= "solid"
%	 stroke-width= "0.5"
%	 p3= "(0.93,-3.44)"
%	 p2= "(0.93,-6.56)"
%	 p1= "(4.06,-6.56)"
%	 closure= "open"
%	 angle-end= "0"
%	 angle-start= "0"
%	 />
%<multicurve fill-style= "none"
%	 stroke-width= "0.5"
%	 points= "(2.5,-5);(2.5,-5);(2.5,-10);(2.5,-10)"
%	 />
%<multicurve fill-style= "none"
%	 stroke-width= "0.5"
%	 points= "(5,-2.5);(5,-2.5);(2.5,-5);(2.5,-5)"
%	 />
%<multicurve fill-style= "none"
%	 stroke-width= "0.5"
%	 points= "(0,-2.5);(0,-2.5);(2.5,-5);(2.5,-5)"
%	 />
%<multicurve fill-style= "none"
%	 stroke-width= "0.5"
%	 points= "(5,0);(5,0);(5,-2.5);(5,-2.5)"
%	 />
%<multicurve fill-style= "none"
%	 stroke-width= "0.5"
%	 points= "(0,0);(0,0);(0,-2.5);(0,-2.5)"
%	 />
%<multicurve fill-style= "none"
%	 stroke-width= "0.5"
%	 points= "(0,5);(0,5);(5,0);(5,0)"
%	 />
%<multicurve fill-style= "none"
%	 stroke-width= "0.5"
%	 points= "(0,0);(0,0);(5,5);(5,5)"
%	 />
%<multicurve fill-style= "none"
%	 stroke-width= "0.5"
%	 points= "(5,7.5);(5,7.5);(5,5);(5,5)"
%	 />
%<multicurve fill-style= "none"
%	 stroke-width= "0.5"
%	 points= "(0,7.5);(0,7.5);(0,5);(0,5)"
%	 />
%</jpic>
%%End JPIC-XML
%PSTricks content-type (pstricks.sty package needed)
%Add \usepackage{pstricks} in the preamble of your LaTeX file
%You can rescale the whole picture (to 80% for instance) by using the command \def\JPicScale{0.8}
\ifx\JPicScale\undefined\def\JPicScale{1}\fi
\psset{unit=\JPicScale mm}
\psset{linewidth=0.3,dotsep=1,hatchwidth=0.3,hatchsep=1.5,shadowsize=1,dimen=middle}
\psset{dotsize=0.7 2.5,dotscale=1 1,fillcolor=black}
\psset{arrowsize=1 2,arrowlength=1,arrowinset=0.25,tbarsize=0.7 5,bracketlength=0.15,rbracketlength=0.15}
\begin{pspicture}(0,0)(5,7.5)
\rput{0}(2.49,-5){\psellipse[linewidth=0.5,fillstyle=solid](0,0)(1.56,-1.56)}
\psline[linewidth=0.5](2.5,-5)(2.5,-10)
\psline[linewidth=0.5](5,-2.5)(2.5,-5)
\psline[linewidth=0.5](0,-2.5)(2.5,-5)
\psline[linewidth=0.5](5,0)(5,-2.5)
\psline[linewidth=0.5](0,0)(0,-2.5)
\psline[linewidth=0.5](0,5)(5,0)
\psline[linewidth=0.5](0,0)(5,5)
\psline[linewidth=0.5](5,7.5)(5,5)
\psline[linewidth=0.5](0,7.5)(0,5)
\end{pspicture}
 \hspace{2em}
\label{eq:comonoid}%\\%[1ex]
\notag
\end{alignat}
\smallskip

\paragraph{State parametrization and updating.} Products $A\otimes B$ denote a space where $A$ and $B$ but do not interfere. In a diagram, they are just parallel strings. Since the product states from the space $X\otimes A$ do not interfere, a transition $g\colon X\otimes A \to B$ can be viewed as $X$-parametrized family $g_{x}\colon A\to B$, as it was viewed in Sec.~\ref{Sec:Intro}. Since the product states from $X\otimes B$ also remain separate, a transition $q\colon X\otimes A\to X\otimes B$ can be viewed as $X$-updating process, as it was also viewed in Sec.~\ref{Sec:Intro}. The corresponding string diagrams are 
\beq\label{eq:gq}
\begin{split}
\newcommand{\fee}{g}
\newcommand{\qee}{q}
\newcommand{\Aee}{\scriptstyle X}
\newcommand{\Bee}{\scriptstyle A}
\newcommand{\Cee}{\scriptstyle B}
\def\JPicScale{.25}
\ifx\JPicScale\undefined\def\JPicScale{1}\fi
\unitlength \JPicScale mm
\begin{picture}(300,120)(0,0)
\linethickness{0.3mm}
\put(80,40){\line(0,1){40}}
\linethickness{0.3mm}
\put(0,40){\line(1,0){80}}
\linethickness{0.3mm}
\put(0,80){\line(1,0){80}}
\linethickness{0.3mm}
\put(60,0){\line(0,1){40}}
\linethickness{0.3mm}
\put(40,80){\line(0,1){40}}
\linethickness{0.3mm}
\put(0,40){\line(0,1){40}}
\linethickness{0.3mm}
\put(20,0){\line(0,1){40}}
\put(45,117.5){\makebox(0,0)[cl]{$\Cee$}}

\put(15,2.5){\makebox(0,0)[cr]{$\Aee$}}

\put(55,2.5){\makebox(0,0)[cr]{$\Bee$}}

\put(40,60){\makebox(0,0)[cc]{$\fee$}}

\linethickness{0.3mm}
\put(300,40){\line(0,1){40}}
\linethickness{0.3mm}
\put(220,40){\line(1,0){80}}
\linethickness{0.3mm}
\put(220,80){\line(1,0){80}}
\linethickness{0.3mm}
\put(280,0){\line(0,1){40}}
\linethickness{0.3mm}
\put(280,80){\line(0,1){40}}
\linethickness{0.3mm}
\put(220,40){\line(0,1){40}}
\linethickness{0.3mm}
\put(240,0){\line(0,1){40}}
\put(285,117.5){\makebox(0,0)[cl]{$\Cee$}}

\put(235,2.5){\makebox(0,0)[cr]{$\Aee$}}

\put(275,2.5){\makebox(0,0)[cr]{$\Bee$}}

\linethickness{0.3mm}
\put(240,80){\line(0,1){40}}
\put(245,117.5){\makebox(0,0)[cl]{$\Aee$}}

\put(260,60){\makebox(0,0)[cc]{$\qee$}}

\end{picture}
 
\end{split}
\eeq

\paragraph{Shape conventions.} While the boxes in \eqref{eq:godement} and \eqref{eq:gq} are rectangular, the cartesian ``boxes'' in \eqref{eq:dataserv} are reduced to black dots. In general, the boxes denoting general transitions can vary in shape, and fixed shapes are used for generic notations. E.g., the interpreters, introduced in \eqref{eq:uev} below, are denoted by trapezoids, and the interpretations, that are fed to them, by triangles. A black dot on a box signals that it is cartesian, i.e. belongs to $\tot\UUU$.

\paragraph{Projections.} Using the cartesian structure from \eqref{eq:dataserv}, a state updating transition $q$ can still be decomposed like before 
\beq
\sta q = \left(X\otimes A\tto q X\times B\tto{\id\otimes \scun} X\right)\quad \out q = \left(X\otimes A\tto q X\otimes B\tto{\scun\otimes \id} B\right)
\eeq
In general, however, although the transitions $u\colon Z\to U$ and $v\colon Z\to V$ can be paired into $<u,v> =(Z\tto\cmn Z\otimes Z \tto{u\otimes v} U\otimes V)$, the pair $<\sta q, \out q>$ may not be equal to $q$ in the universe $\UUU$, unless it happens to be cloneable, in the sense that it commutes with $\cmn$.

\section{Universal language}\label{Sec:moncom}
% !TEX root = 00-wollic.tex

A theory of theories, such as the categorical theory of sketches, is a theory. Category theory is also a theory and functorial semantics provides a categorical theory of reference models. The theory of state spaces from Sec.~\ref{Sec:state} can thus be formalized and presented as a state space in the category $\UUU$. The theory of state spaces from Sec.~\ref{Sec:state} can thus be formalized into a sketch with a reference model and presented as a state space in the category $\UUU$. The theory of state transitions from Sec.~\ref{Sec:transition} is another sketch, and with another reference model it is also a  state space in $\UUU$. Call it $\DP$. The fact that the states in $\DP$ correspond to the transitions in $\UUU$ means that it satisfies a parametrized version of \eqref{eq:univ}. It is a universal language for $\UUU$. Its interpreters follow from its definition, as the models of the theory of transitions. Since there is no room here to spell out the details of a theory of transitions and show that the correspondence of its cartesian models and the transitions in $\UUU$ equips $\DP$ with all interpreters, we  postulate the existence of the interpreters by the following definition. 

\begin{definition}\label{Def:interpreter}
An \emph{universal interpreter}\/ for state spaces $A,B$ is a transition $\{\}\colon \DP\otimes A\to B$ in $\UUU$ which is universal for all parametric families of transitions from $A$ to $B$. This means that for any state space $X$ and any transition $g\in \UUU(X\otimes A, B)$ there is an interpretation $G\in \tot\UUU(X,\DP)$ with
\beq\label{eq:uev}
\begin{split}
\newcommand{\Fee}{\scriptstyle G}
\newcommand{\fee}{g}
\newcommand{\Aee}{\scriptstyle X}
\newcommand{\Bee}{\scriptstyle A}
\newcommand{\Cee}{\scriptstyle B}
\newcommand{\Code}{\scriptstyle \PPp}
\newcommand{\Univ}{\mbox{\large$\{\}$}}
\newcommand{\Dott}{\mbox{\Large$\bullet$}}
\def\JPicScale{.33}
\ifx\JPicScale\undefined\def\JPicScale{1}\fi
\unitlength \JPicScale mm
\begin{picture}(220,120)(0,0)
\linethickness{0.35mm}
\put(80,40){\line(0,1){40}}
\linethickness{0.35mm}
\put(0,40){\line(1,0){80}}
\linethickness{0.35mm}
\put(0,80){\line(1,0){80}}
\linethickness{0.35mm}
\put(60,0){\line(0,1){40}}
\linethickness{0.35mm}
\put(40,80){\line(0,1){40}}
\linethickness{0.35mm}
\put(0,40){\line(0,1){40}}
\linethickness{0.35mm}
\put(20,0){\line(0,1){40}}
\linethickness{0.35mm}
\put(220,60){\line(0,1){40}}
\linethickness{0.35mm}
\put(140,20){\line(1,0){40}}
\linethickness{0.35mm}
\put(140,100){\line(1,0){80}}
\linethickness{0.35mm}
\put(200,0){\line(0,1){60}}
\linethickness{0.35mm}
\put(180,100){\line(0,1){20}}
\linethickness{0.35mm}
\put(140,20){\line(0,1){40}}
\linethickness{0.35mm}
\put(160,0){\line(0,1){20}}
\linethickness{0.35mm}
\put(180,60){\line(1,0){40}}
\linethickness{0.35mm}
\multiput(140,100)(0.12,-0.12){333}{\line(1,0){0.12}}
\linethickness{0.35mm}
\multiput(140,60)(0.12,-0.12){333}{\line(1,0){0.12}}
\linethickness{0.7mm}
\put(160,40){\line(0,1){40}}
\put(190,80){\makebox(0,0)[cc]{$\Univ$}}

\put(110,60){\makebox(0,0)[cc]{\EQLS}}

\put(156.25,65){\makebox(0,0)[cr]{$\Code$}}

\put(185,117.5){\makebox(0,0)[cl]{$\Cee$}}

\put(45,117.5){\makebox(0,0)[cl]{$\Cee$}}

\put(15,2.5){\makebox(0,0)[cr]{$\Aee$}}

\put(55,2.5){\makebox(0,0)[cr]{$\Bee$}}

\put(40,60){\makebox(0,0)[cc]{$\fee$}}

\put(150,31.25){\makebox(0,0)[cc]{$\Fee$}}

\put(155,2.5){\makebox(0,0)[cr]{$\Aee$}}

\put(195,2.5){\makebox(0,0)[cr]{$\Bee$}}

\put(160,40){\makebox(0,0)[cc]{$\Dott$}}

\end{picture}
 
\end{split}
\eeq
\end{definition}

On one hand, a universal interpreter is universal for parametric families. On the other hand, it is a parametric family itself. It is thus capable of interpreting itself.  This capability of self-reflection was crucial for G\"odel's incompleteness construction.  This capability is embodied in the \emph{specializers}, which are derived  directly from Def~\ref{Def:interpreter}. 

\begin{lemma}\label{prop:pev}
For any $X, A, B$ there is an interpretation $\prtial \in \tot\UUU(\DP\times X, \DP)$ which specializes from a given $X\otimes A$-interpreter to an $A$-interpreter, in the sense
\beq\label{eq:pev}
\begin{split}
\newcommand{\Fee}{\prtial}
\newcommand{\fee}{\mbox{\large$\{\}$}}
\newcommand{\Aee}{\scriptstyle X}
\newcommand{\Bee}{\scriptstyle A}
\newcommand{\Cee}{\scriptstyle B}
\newcommand{\Code}{\scriptstyle \DP}
\newcommand{\Univ}{\mbox{\large$\{\}$}}
\newcommand{\Dott}{\mbox{\LARGE$\bullet$}}
\def\JPicScale{.33}
\ifx\JPicScale\undefined\def\JPicScale{1}\fi
\unitlength \JPicScale mm
\begin{picture}(280,120)(0,0)
\linethickness{0.3mm}
\put(120,40){\line(0,1){40}}
\linethickness{0.3mm}
\put(40,40){\line(1,0){80}}
\linethickness{0.3mm}
\put(0,80){\line(1,0){120}}
\linethickness{0.3mm}
\put(100,0){\line(0,1){40}}
\linethickness{0.3mm}
\put(80,80){\line(0,1){40}}
\linethickness{0.3mm}
\multiput(0,80)(0.12,-0.12){333}{\line(1,0){0.12}}
\linethickness{0.3mm}
\put(60,0){\line(0,1){40}}
\linethickness{0.3mm}
\put(280,60){\line(0,1){40}}
\linethickness{0.3mm}
\put(200,20){\line(1,0){40}}
\linethickness{0.3mm}
\put(200,100){\line(1,0){80}}
\linethickness{0.3mm}
\put(260,0){\line(0,1){60}}
\linethickness{0.3mm}
\put(240,100){\line(0,1){20}}
\linethickness{0.3mm}
\multiput(160,60)(0.12,-0.12){333}{\line(1,0){0.12}}
\linethickness{0.3mm}
\put(220,0){\line(0,1){20}}
\linethickness{0.3mm}
\put(240,60){\line(1,0){40}}
\linethickness{0.3mm}
\multiput(200,100)(0.12,-0.12){333}{\line(1,0){0.12}}
\linethickness{0.3mm}
\multiput(200,60)(0.12,-0.12){333}{\line(1,0){0.12}}
\linethickness{0.7mm}
\put(220,40){\line(0,1){40}}
\put(250,80){\makebox(0,0)[cc]{$\Univ$}}

\put(140,60){\makebox(0,0)[cc]{\EQLS}}

\put(216.25,65){\makebox(0,0)[cr]{$\Code$}}

\put(245,117.5){\makebox(0,0)[cl]{$\Cee$}}

\put(85,117.5){\makebox(0,0)[cl]{$\Cee$}}

\put(55,2.5){\makebox(0,0)[cr]{$\Aee$}}

\put(95,2.5){\makebox(0,0)[cr]{$\Bee$}}

\put(80,60){\makebox(0,0)[cc]{$\fee$}}

\put(200,40){\makebox(0,0)[cc]{$\Fee$}}

\put(215,2.5){\makebox(0,0)[cr]{$\Aee$}}

\put(255,2.5){\makebox(0,0)[cr]{$\Bee$}}

\put(220,40){\makebox(0,0)[cc]{$\Dott$}}

\linethickness{0.3mm}
\put(20,0){\line(0,1){60}}
\linethickness{0.3mm}
\put(160,60){\line(1,0){40}}
\linethickness{0.3mm}
\put(180,0){\line(0,1){40}}
\put(175,2.5){\makebox(0,0)[cr]{$\Code$}}

\put(15,2.5){\makebox(0,0)[cr]{$\Code$}}

\end{picture}
 
\end{split}
\eeq
\end{lemma}

\paragraph{Hoare logic of interpreters and specializers.} If interpreters are presented as Hoare triples in the form $(X\otimes A)\uev G B$, and if $X\!\pev G$ denotes a specialization of $G$ to $X$ as above, then \eqref{eq:pev} can be written as the invertible Hoare rule
\[\prooftree
(X\otimes A)\uev G B
\Justifies
A\uev{X\!\pev G} B
\endprooftree\]   

\paragraph{Explanations.} Interpretations  (in the sense of Def.~\ref{Def:interpretable}) of arbitrary states from some space $X$ along $G\in \tot\UUU(X,\DP)$ in a universal language $\DP$ can be construed as \emph{explanations}. If $\DP$ is a programming language, they are programs. The idea that explaining a process means programming a computation has been pursued in theory of science from various directions \cite[and references therein]{Osherson:sci-inquiry}. A universal language $\DP$ is thus a universal space of explanations. The idea of programming languages as universal state spaces is pursued in \cite[Ch.~7]{PavlovicD:MonCom}. Just like any universal programming language makes every computation programmable, any universal language from Def.~\ref{Def:interpreter} makes any observable transition explainable. What we cannot explain, we cannot recognize, and therefore we cannot observe. But it gets funny when we take into account how our explanations influence our observations, and how our current explanations can be made to steer future observations. This is sketched in the next two sections.

\section{Self-explanations}\label{Sec:self-fulfill}
% !TEX root = 00-wollic.tex

When a state change depends on our explanations, then we can find an explanation consistent with its own impact: the state changes the way the explanation predicts. More precisely, if a family of transitions in the form $t\colon \DP\otimes X\otimes A\to B$, then the predictions $t_{\ell x}$ can be steered by varying the explanations $\ell$ for every $x$ until a family of explanations $\enco t \colon X\to \DP$ is found, which is self-confirming at all states $x$, i.e. it satisfies $t(\enco t_{x}, x, a) = \uev{\enco t_{x}} a$. 

\begin{proposition}
For any belief transition $t \in \UUU(\DP\otimes X\otimes A, B)$ there is an explanation $\enco t\in \tot \UUU(X,\DP)$ such that
\beq\label{eq:fund}
\begin{split}
\newcommand{\DOTT}{\mbox{\Large$\bullet$}}
\newcommand{\Aah}{\scriptscriptstyle A}
\newcommand{\Xah}{\scriptscriptstyle X}
\newcommand{\grr}{t}
\newcommand{\Bah}{\scriptscriptstyle B}
\newcommand{\GRR}{\scriptstyle \enco t}
\newcommand{\UK}{\mbox{\large$\{\}$}}
\def\JPicScale{.55}
\ifx\JPicScale\undefined\def\JPicScale{1}\fi
\unitlength \JPicScale mm
\begin{picture}(115,90)(0,0)
\linethickness{0.3mm}
\put(5,70){\line(1,0){50}}
\linethickness{0.3mm}
\put(55,50){\line(0,1){20}}
\linethickness{0.3mm}
\put(5,50){\line(1,0){50}}
\linethickness{0.3mm}
\put(5,50){\line(0,1){20}}
\linethickness{0.3mm}
\put(30,70){\line(0,1){20}}
\linethickness{0.65mm}
\put(10,35){\line(0,1){15}}
\linethickness{0.3mm}
\put(30,20){\line(0,1){30}}
\linethickness{0.3mm}
\multiput(0,45)(0.12,-0.12){167}{\line(1,0){0.12}}
\linethickness{0.3mm}
\put(0,25){\line(1,0){20}}
\linethickness{0.3mm}
\put(0,25){\line(0,1){20}}
\linethickness{0.3mm}
\put(75,70){\line(1,0){40}}
\linethickness{0.3mm}
\put(115,50){\line(0,1){20}}
\linethickness{0.3mm}
\put(95,50){\line(1,0){20}}
\linethickness{0.3mm}
\multiput(75,70)(0.12,-0.12){167}{\line(1,0){0.12}}
\linethickness{0.3mm}
\put(100,70){\line(0,1){20}}
\linethickness{0.3mm}
\put(10,20){\line(0,1){5}}
\linethickness{0.3mm}
\multiput(75,45)(0.12,-0.12){167}{\line(1,0){0.12}}
\linethickness{0.3mm}
\put(75,25){\line(1,0){20}}
\linethickness{0.3mm}
\put(75,25){\line(0,1){20}}
\linethickness{0.65mm}
\put(85,35){\line(0,1){25}}
\put(10,35){\makebox(0,0)[cc]{\DOTT}}

\put(85,35){\makebox(0,0)[cc]{\DOTT}}

\put(46.25,0){\makebox(0,0)[cr]{$\Aah$}}

\put(107.5,0){\makebox(0,0)[cr]{$\Aah$}}

\put(30,60){\makebox(0,0)[cc]{$\grr$}}

\put(32.5,90){\makebox(0,0)[cl]{$\Bah$}}

\put(102.5,90){\makebox(0,0)[cl]{$\Bah$}}

\put(100,60){\makebox(0,0)[cc]{$\UK$}}

\put(65,60){\makebox(0,0)[cc]{\EQLS}}

\put(5,30){\makebox(0,0)[cc]{$\GRR$}}

\put(80,30){\makebox(0,0)[cc]{$\GRR$}}

\linethickness{0.3mm}
\put(50,0){\line(0,1){50}}
\linethickness{0.3mm}
\put(110,0){\line(0,1){50}}
\linethickness{0.3mm}
\multiput(10,20)(0.12,-0.12){83}{\line(1,0){0.12}}
\linethickness{0.3mm}
\multiput(20,10)(0.12,0.12){83}{\line(1,0){0.12}}
\linethickness{0.3mm}
\put(20,0){\line(0,1){10}}
\linethickness{0.3mm}
\put(85,0){\line(0,1){25}}
\put(82.5,0){\makebox(0,0)[cr]{$\Xah$}}

\put(17.5,0){\makebox(0,0)[cr]{$\Xah$}}

\put(20,10){\makebox(0,0)[cc]{\DOTT}}

\end{picture}

\end{split}
\eeq 
\end{proposition}

\bpr
Let $T\in \tot\UUU(X,\DP)$ be an explanation of the transition on the left in \eqref{eq:fund}. 
\beq
\begin{split}
\newcommand{\DOTT}{\mbox{\Large$\bullet$}}
\newcommand{\Aah}{\scriptscriptstyle A}
\newcommand{\Xah}{\scriptscriptstyle X}
\newcommand{\grr}{t}
\newcommand{\Bah}{\scriptscriptstyle B}
\newcommand{\Grr}{\scriptstyle T}
\newcommand{\UK}{\universal}
\newcommand{\PK}{\prtial}
\def\JPicScale{.45}
\ifx\JPicScale\undefined\def\JPicScale{1}\fi
\unitlength \JPicScale mm
\begin{picture}(155,120)(0,0)
\linethickness{0.3mm}
\put(25,100){\line(1,0){50}}
\linethickness{0.3mm}
\put(75,80){\line(0,1){20}}
\linethickness{0.3mm}
\put(25,80){\line(1,0){50}}
\linethickness{0.3mm}
\put(25,80){\line(0,1){20}}
\linethickness{0.3mm}
\put(50,100){\line(0,1){20}}
\linethickness{0.65mm}
\put(30,60){\line(0,1){20}}
\linethickness{0.3mm}
\put(70,0){\line(0,1){80}}
\linethickness{0.3mm}
\multiput(20,70)(0.12,-0.12){167}{\line(1,0){0.12}}
\linethickness{0.3mm}
\put(20,50){\line(1,0){20}}
\linethickness{0.3mm}
\multiput(0,70)(0.12,-0.12){167}{\line(1,0){0.12}}
\linethickness{0.3mm}
\put(0,70){\line(1,0){20}}
\linethickness{0.65mm}
\put(30,30){\line(0,1){20}}
\linethickness{0.3mm}
\put(100,100){\line(1,0){55}}
\linethickness{0.3mm}
\put(155,80){\line(0,1){20}}
\linethickness{0.3mm}
\put(120,80){\line(1,0){35}}
\linethickness{0.3mm}
\multiput(100,100)(0.12,-0.12){167}{\line(1,0){0.12}}
\linethickness{0.3mm}
\put(140,100){\line(0,1){20}}
\linethickness{0.65mm}
\put(130,0){\line(0,1){80}}
\linethickness{0.3mm}
\put(150,0){\line(0,1){80}}
\linethickness{0.3mm}
\multiput(100,70)(0.12,-0.12){167}{\line(1,0){0.12}}
\linethickness{0.3mm}
\put(100,50){\line(1,0){20}}
\linethickness{0.3mm}
\put(100,50){\line(0,1){20}}
\linethickness{0.65mm}
\put(110,60){\line(0,1){30}}
\linethickness{0.65mm}
\put(10,50){\line(0,1){10}}
\put(30,30){\makebox(0,0)[cc]{\DOTT}}

\put(30,60){\makebox(0,0)[cc]{\DOTT}}

\put(110,60){\makebox(0,0)[cc]{\DOTT}}

\put(67.5,0){\makebox(0,0)[cr]{$\Aah$}}

\put(147.5,0){\makebox(0,0)[cr]{$\Aah$}}

\put(20,60){\makebox(0,0)[cc]{$\PK$}}

\put(50,90){\makebox(0,0)[cc]{$\grr$}}

\put(52.5,120){\makebox(0,0)[cl]{$\Bah$}}

\put(142.5,120){\makebox(0,0)[cl]{$\Bah$}}

\put(140,90){\makebox(0,0)[cc]{$\UK$}}

\put(90,90){\makebox(0,0)[cc]{\EQLS}}

\put(105,55){\makebox(0,0)[cc]{$\Grr$}}

\linethickness{0.3mm}
\put(50,30){\line(0,1){50}}
\linethickness{0.3mm}
\multiput(30,10)(0.12,0.12){167}{\line(1,0){0.12}}
\linethickness{0.3mm}
\put(30,0){\line(0,1){10}}
\linethickness{0.65mm}
\put(50,0){\line(0,1){10}}
\linethickness{0.65mm}
\multiput(10,50)(0.12,-0.12){333}{\line(1,0){0.12}}
\linethickness{0.3mm}
\put(110,0){\line(0,1){50}}
\put(107.5,0){\makebox(0,0)[cr]{$\Xah$}}

\put(27.5,0){\makebox(0,0)[cr]{$\Xah$}}

\end{picture}

\end{split}
\eeq
$H$ exists by Def.~\ref{Def:interpreter}. Then $\enco t_{x} =\pev{Tx}$ is self-confirming, because
\beq
\begin{split}
\newcommand{\DOTT}{\mbox{\Large$\bullet$}}
\newcommand{\Aah}{\scriptscriptstyle A}
\newcommand{\Xah}{\scriptscriptstyle X}
\newcommand{\grr}{t}
\newcommand{\Bah}{\scriptscriptstyle B}
\newcommand{\Grr}{\scriptstyle T}
\newcommand{\Psee}{\enco t}
\newcommand{\UK}{\universal}
\newcommand{\PK}{\prtial}
\def\JPicScale{.45}
\ifx\JPicScale\undefined\def\JPicScale{1}\fi
\unitlength \JPicScale mm
\begin{picture}(320,140)(0,0)
\linethickness{0.3mm}
\put(30,120){\line(1,0){50}}
\linethickness{0.3mm}
\put(80,100){\line(0,1){20}}
\linethickness{0.3mm}
\put(30,100){\line(1,0){50}}
\linethickness{0.3mm}
\put(30,100){\line(0,1){20}}
\linethickness{0.3mm}
\put(55,120){\line(0,1){20}}
\linethickness{0.65mm}
\put(35,80){\line(0,1){20}}
\linethickness{0.3mm}
\put(75,0){\line(0,1){100}}
\linethickness{0.3mm}
\multiput(25,90)(0.12,-0.12){167}{\line(1,0){0.12}}
\linethickness{0.3mm}
\put(25,70){\line(1,0){20}}
\linethickness{0.3mm}
\multiput(5,90)(0.12,-0.12){167}{\line(1,0){0.12}}
\linethickness{0.3mm}
\put(5,90){\line(1,0){20}}
\linethickness{0.65mm}
\put(35,45){\line(0,1){25}}
\linethickness{0.3mm}
\put(105,120){\line(1,0){55}}
\linethickness{0.3mm}
\put(160,100){\line(0,1){20}}
\linethickness{0.3mm}
\put(125,100){\line(1,0){35}}
\linethickness{0.3mm}
\multiput(105,120)(0.12,-0.12){167}{\line(1,0){0.12}}
\linethickness{0.3mm}
\put(145,120){\line(0,1){20}}
\linethickness{0.65mm}
\put(135,45){\line(0,1){55}}
\linethickness{0.3mm}
\put(155,0){\line(0,1){100}}
\linethickness{0.3mm}
\multiput(105,90)(0.12,-0.12){167}{\line(1,0){0.12}}
\linethickness{0.3mm}
\put(105,70){\line(1,0){20}}
\linethickness{0.3mm}
\put(105,70){\line(0,1){20}}
\linethickness{0.65mm}
\put(115,80){\line(0,1){30}}
\linethickness{0.65mm}
\put(15,72.5){\line(0,1){7.5}}
\put(35,52.5){\makebox(0,0)[cc]{\DOTT}}

\put(35,80){\makebox(0,0)[cc]{\DOTT}}

\put(115,80){\makebox(0,0)[cc]{\DOTT}}

\put(72.5,0){\makebox(0,0)[cr]{$\Aah$}}

\put(152.5,0){\makebox(0,0)[cr]{$\Aah$}}

\put(25,80){\makebox(0,0)[cc]{$\PK$}}

\put(55,110){\makebox(0,0)[cc]{$\grr$}}

\put(57.5,140){\makebox(0,0)[cl]{$\Bah$}}

\put(147.5,140){\makebox(0,0)[cl]{$\Bah$}}

\put(145,110){\makebox(0,0)[cc]{$\UK$}}

\put(95,110){\makebox(0,0)[cc]{\EQLS}}

\put(110,75){\makebox(0,0)[cc]{$\Grr$}}

\linethickness{0.3mm}
\put(55,30){\line(0,1){70}}
\linethickness{0.3mm}
\put(35,0){\line(0,1){35}}
\linethickness{0.65mm}
\multiput(15,72.5)(0.12,-0.12){167}{\line(1,0){0.12}}
\linethickness{0.3mm}
\put(115,0){\line(0,1){70}}
\put(112.5,0){\makebox(0,0)[cr]{$\Xah$}}

\put(32.5,0){\makebox(0,0)[cr]{$\Xah$}}

\linethickness{0.3mm}
\put(-0,95){\line(1,0){25}}
\linethickness{0.3mm}
\multiput(25,95)(0.12,-0.12){208}{\line(1,0){0.12}}
\linethickness{0.3mm}
\put(50,30){\line(0,1){40}}
\linethickness{0.3mm}
\put(0,30){\line(1,0){50}}
\linethickness{0.3mm}
\multiput(25,55)(0.12,-0.12){167}{\line(1,0){0.12}}
\linethickness{0.3mm}
\put(25,35){\line(1,0){20}}
\linethickness{0.3mm}
\put(25,35){\line(0,1){20}}
\put(35,45){\makebox(0,0)[cc]{\DOTT}}

\put(30,40){\makebox(0,0)[cc]{$\Grr$}}

\linethickness{0.3mm}
\multiput(35,10)(0.12,0.12){167}{\line(1,0){0.12}}
\put(35,10){\makebox(0,0)[cc]{\DOTT}}

\linethickness{0.3mm}
\put(0,30){\line(0,1){65}}
\linethickness{0.3mm}
\multiput(125,55)(0.12,-0.12){167}{\line(1,0){0.12}}
\linethickness{0.3mm}
\put(125,35){\line(1,0){20}}
\linethickness{0.3mm}
\put(125,35){\line(0,1){20}}
\put(135,45){\makebox(0,0)[cc]{\DOTT}}

\put(130,40){\makebox(0,0)[cc]{$\Grr$}}

\linethickness{0.3mm}
\multiput(115,10)(0.12,0.12){167}{\line(1,0){0.12}}
\linethickness{0.3mm}
\put(135,30){\line(0,1){5}}
\linethickness{0.3mm}
\put(185,120){\line(1,0){55}}
\linethickness{0.3mm}
\put(240,100){\line(0,1){20}}
\linethickness{0.3mm}
\put(205,100){\line(1,0){35}}
\linethickness{0.3mm}
\multiput(185,120)(0.12,-0.12){167}{\line(1,0){0.12}}
\linethickness{0.3mm}
\put(225,120){\line(0,1){20}}
\linethickness{0.65mm}
\put(215,45){\line(0,1){55}}
\linethickness{0.3mm}
\put(235,0){\line(0,1){100}}
\put(232.5,0){\makebox(0,0)[cr]{$\Aah$}}

\put(227.5,140){\makebox(0,0)[cl]{$\Bah$}}

\put(225,110){\makebox(0,0)[cc]{$\UK$}}

\put(175,110){\makebox(0,0)[cc]{\EQLS}}

\put(212.5,0){\makebox(0,0)[cr]{$\Xah$}}

\linethickness{0.3mm}
\multiput(205,55)(0.12,-0.12){167}{\line(1,0){0.12}}
\linethickness{0.3mm}
\put(205,35){\line(1,0){20}}
\linethickness{0.3mm}
\put(205,35){\line(0,1){20}}
\put(215,45){\makebox(0,0)[cc]{\DOTT}}

\put(210,40){\makebox(0,0)[cc]{$\Grr$}}

\linethickness{0.3mm}
\put(215,0){\line(0,1){35}}
\linethickness{0.3mm}
\put(285,120){\line(1,0){35}}
\linethickness{0.3mm}
\put(320,100){\line(0,1){20}}
\linethickness{0.3mm}
\put(305,100){\line(1,0){15}}
\linethickness{0.3mm}
\multiput(285,120)(0.12,-0.12){167}{\line(1,0){0.12}}
\linethickness{0.3mm}
\put(305,120){\line(0,1){20}}
\linethickness{0.3mm}
\put(315,0){\line(0,1){100}}
\put(312.5,0){\makebox(0,0)[cr]{$\Aah$}}

\put(307.5,140){\makebox(0,0)[cl]{$\Bah$}}

\put(310,110){\makebox(0,0)[cc]{$\UK$}}

\put(255,110){\makebox(0,0)[cc]{\EQLS}}

\linethickness{0.65mm}
\put(295,80){\line(0,1){30}}
\linethickness{0.3mm}
\multiput(285,90)(0.12,-0.12){167}{\line(1,0){0.12}}
\linethickness{0.3mm}
\put(285,70){\line(1,0){20}}
\linethickness{0.3mm}
\multiput(265,90)(0.12,-0.12){167}{\line(1,0){0.12}}
\linethickness{0.3mm}
\put(265,90){\line(1,0){20}}
\linethickness{0.65mm}
\put(295,45){\line(0,1){25}}
\linethickness{0.65mm}
\put(275,72.5){\line(0,1){7.5}}
\put(295,52.5){\makebox(0,0)[cc]{\DOTT}}

\put(295,80){\makebox(0,0)[cc]{\DOTT}}

\put(285,80){\makebox(0,0)[cc]{$\PK$}}

\linethickness{0.3mm}
\put(295,0){\line(0,1){35}}
\linethickness{0.65mm}
\multiput(275,72.5)(0.12,-0.12){167}{\line(1,0){0.12}}
\put(292.5,0){\makebox(0,0)[cr]{$\Xah$}}

\linethickness{0.3mm}
\put(260,95){\line(1,0){25}}
\linethickness{0.3mm}
\multiput(285,95)(0.12,-0.12){208}{\line(1,0){0.12}}
\linethickness{0.3mm}
\put(310,30){\line(0,1){40}}
\linethickness{0.3mm}
\put(260,30){\line(1,0){50}}
\linethickness{0.3mm}
\multiput(285,55)(0.12,-0.12){167}{\line(1,0){0.12}}
\linethickness{0.3mm}
\put(285,35){\line(1,0){20}}
\linethickness{0.3mm}
\put(285,35){\line(0,1){20}}
\put(295,45){\makebox(0,0)[cc]{\DOTT}}

\put(290,40){\makebox(0,0)[cc]{$\Grr$}}

\linethickness{0.3mm}
\put(260,30){\line(0,1){65}}
\linethickness{0.65mm}
\put(195,72.5){\line(0,1){37.5}}
\put(215,52.5){\makebox(0,0)[cc]{\DOTT}}

\linethickness{0.65mm}
\multiput(195,72.5)(0.12,-0.12){167}{\line(1,0){0.12}}
\put(10,50){\makebox(0,0)[cc]{$\Psee$}}

\put(270,50){\makebox(0,0)[cc]{$\Psee$}}

\put(115,10){\makebox(0,0)[cc]{\DOTT}}

\end{picture}

\end{split}
\eeq
\epr

\section{Unfalsifiable explanations}\label{Sec:unfalse}
% !TEX root = 00-wollic.tex

A transition in the form $q\colon X\otimes A\to X\otimes B$ updates the state $x$ on input $a$ to a state $x'=\sta q_{x}(a)$ in $X$ and moreover produces an output $b=\out q_{x}(a)$ in $B$. A correct explanation $\ana q\colon X\to \DP$ of the process $q$ must  correctly predict the next state and the output. The predictions are extracted from an explanation by the interpreter $\uuniversal$. In this case, the predictions of an explanation   $\ana q$ of the process $q$ at a state $x$ and on an input $a$ will be in the form $\uev{\ana q_{x}}(a)$ in $X\otimes B$. A correct prediction of the output $b=\out q_{x}(a)$ is simply $\out{\uev{\ana q_{x}}}(a) = b$. However, the external state $x'=\sta q_{x}(a)$ may not be directly observable. It is \emph{believed}\/ to be explained by $\ana q_{x'}$. A correct prediction of the next state is thus a correct prediction of its explanation $\sta{\uev{\ana q_{x}}}(a) = \ana q_{x'}$. At each state $x$, the explanation $\ana q_{x}$ is required to anticipate the explanations $\ana q_{x'}$ of all future states and be consistent with them. If the explanation $\ana q _{x'}$ at a future state $x'=\sta q_{x}(a)$ is found to be inconsistent with the explanation $\sta{\uev{\ana q_{x}}}(a)$, then the explanations $\ana q$ of the process $q$ have been proven false. This is the standard process of testing explanations. Our claim is, however, that a universal language allows constructing \emph{testable but unfalsifiable explanations}, that remain consistent at all future states. This persistent consistency can be viewed as a dynamic form of completeness. It is achieved by predicting the state updates of the given process $q$ and anticipating their explanations, as in the following construction.

\begin{proposition}
For any process $q \in \UUU(X\otimes A, X\otimes B)$ there is an explanation $\ana q\in \tot \UUU(X,\DP)$ which maintains consistency of all future explanations:
\beq
\begin{split}
\newcommand{\Aee}{\scriptstyle A}
\newcommand{\geee}{q}
\newcommand{\Beee}{\scriptstyle B}
\newcommand{\XH}{\scriptstyle X}
\newcommand{\PPh}{\scriptstyle \DP}
\newcommand{\Univv}{\universal}
\newcommand{\dottt}{\mbox{\Large$\bullet$}}
\newcommand{\Mhh}{\scriptstyle\ana q}
\def\JPicScale{.55}
\ifx\JPicScale\undefined\def\JPicScale{1}\fi
\unitlength \JPicScale mm
\begin{picture}(100,80)(0,0)
\linethickness{0.35mm}
\put(60,40){\line(1,0){40}}
\linethickness{0.35mm}
\put(100,20){\line(0,1){20}}
\linethickness{0.75mm}
\put(70,40){\line(0,1){40}}
\linethickness{0.35mm}
\put(80,20){\line(1,0){20}}
\put(50,30){\makebox(0,0)[cc]{\EQLS}}

\linethickness{0.35mm}
\multiput(60,40)(0.12,-0.12){167}{\line(1,0){0.12}}
\linethickness{0.35mm}
\put(90,0){\line(0,1){20}}
\linethickness{0.35mm}
\put(90,40){\line(0,1){40}}
\put(73.75,80){\makebox(0,0)[cl]{$\PPh$}}

\put(67.5,0){\makebox(0,0)[cr]{$\XH$}}

\linethickness{0.35mm}
\put(70,0){\line(0,1){10}}
\linethickness{0.75mm}
\put(70,20){\line(0,1){10}}
\put(65,15){\makebox(0,0)[cc]{$\Mhh$}}

\put(87.5,0){\makebox(0,0)[cr]{$\Aee$}}

\put(93.75,80){\makebox(0,0)[cl]{$\Beee$}}

\put(20,30){\makebox(0,0)[cc]{$\geee$}}

\linethickness{0.35mm}
\multiput(60,30)(0.12,-0.12){167}{\line(1,0){0.12}}
\linethickness{0.35mm}
\put(60,10){\line(1,0){20}}
\linethickness{0.35mm}
\put(60,10){\line(0,1){20}}
\linethickness{0.35mm}
\put(0,40){\line(1,0){40}}
\linethickness{0.35mm}
\put(40,20){\line(0,1){20}}
\linethickness{0.35mm}
\put(10,40){\line(0,1){10}}
\linethickness{0.35mm}
\put(0,20){\line(1,0){40}}
\linethickness{0.35mm}
\put(30,0){\line(0,1){20}}
\linethickness{0.35mm}
\put(30,40){\line(0,1){40}}
\put(12.5,80){\makebox(0,0)[cl]{$\PPh$}}

\put(7.5,0){\makebox(0,0)[cr]{$\XH$}}

\linethickness{0.35mm}
\put(10,0){\line(0,1){20}}
\linethickness{0.75mm}
\put(10,60){\line(0,1){20}}
\put(27.5,0){\makebox(0,0)[cr]{$\Aee$}}

\put(33.75,80){\makebox(0,0)[cl]{$\Beee$}}

\linethickness{0.35mm}
\multiput(0,70)(0.12,-0.12){167}{\line(1,0){0.12}}
\linethickness{0.35mm}
\put(0,50){\line(1,0){20}}
\linethickness{0.35mm}
\put(0,20){\line(0,1){20}}
\linethickness{0.35mm}
\put(0,50){\line(0,1){20}}
\put(10,60){\makebox(0,0)[cc]{$\dottt$}}

\put(70,20){\makebox(0,0)[cc]{$\dottt$}}

\put(5,55){\makebox(0,0)[cc]{$\Mhh$}}

\put(85,30){\makebox(0,0)[cc]{$\Univv$}}

\end{picture}

\end{split}
\eeq 
\end{proposition}

\bpr
Set $\ana q = \pev Q$ where $Q$ is an explanation of the belief transition $q$ postcomposed with a specialization over the state space $X$ of updates:
\beq
\begin{split}
\newcommand{\Aee}{}%{\scriptstyle A}
\newcommand{\geee}{q}
\newcommand{\Beee}{}%{\scriptstyle B}
\newcommand{\XH}{}%{\scriptstyle X}
\newcommand{\PPh}{}%{\scriptstyle \DP}
\newcommand{\Univv}{\universal}
\newcommand{\dottt}{\mbox{\Large$\bullet$}}
\newcommand{\Mhh}{\scriptstyle Q}
\newcommand{\prtls}{\prtial}
\def\JPicScale{.55}
\ifx\JPicScale\undefined\def\JPicScale{1}\fi
\unitlength \JPicScale mm
\begin{picture}(220,80)(0,0)
\linethickness{0.35mm}
\put(80,40){\line(1,0){60}}
\linethickness{0.35mm}
\put(140,20){\line(0,1){20}}
\linethickness{0.75mm}
\put(110,40){\line(0,1){40}}
\linethickness{0.35mm}
\put(100,20){\line(1,0){40}}
\put(70,30){\makebox(0,0)[cc]{\EQLS}}

\linethickness{0.35mm}
\multiput(80,40)(0.12,-0.12){167}{\line(1,0){0.12}}
\linethickness{0.35mm}
\put(130,0){\line(0,1){20}}
\linethickness{0.35mm}
\put(130,40){\line(0,1){40}}
\put(113.75,80){\makebox(0,0)[cl]{$\PPh$}}

\put(107.5,0){\makebox(0,0)[cr]{$\XH$}}

\linethickness{0.35mm}
\put(110,0){\line(0,1){20}}
\linethickness{0.75mm}
\put(90,10){\line(0,1){20}}
\put(85,5){\makebox(0,0)[cc]{$\Mhh$}}

\put(127.5,0){\makebox(0,0)[cr]{$\Aee$}}

\put(133.75,80){\makebox(0,0)[cl]{$\Beee$}}

\put(40,30){\makebox(0,0)[cc]{$\geee$}}

\linethickness{0.35mm}
\multiput(80,20)(0.12,-0.12){167}{\line(1,0){0.12}}
\linethickness{0.35mm}
\put(80,0){\line(1,0){20}}
\linethickness{0.35mm}
\put(80,0){\line(0,1){20}}
\linethickness{0.35mm}
\put(20,40){\line(1,0){40}}
\linethickness{0.35mm}
\put(60,20){\line(0,1){20}}
\linethickness{0.35mm}
\put(30,40){\line(0,1){10}}
\linethickness{0.35mm}
\put(20,20){\line(1,0){40}}
\linethickness{0.35mm}
\put(50,0){\line(0,1){20}}
\linethickness{0.35mm}
\put(50,40){\line(0,1){40}}
\put(32.5,80){\makebox(0,0)[cl]{$\PPh$}}

\put(27.5,0){\makebox(0,0)[cr]{$\XH$}}

\linethickness{0.35mm}
\put(30,0){\line(0,1){20}}
\linethickness{0.75mm}
\put(30,60){\line(0,1){20}}
\put(47.5,0){\makebox(0,0)[cr]{$\Aee$}}

\put(53.75,80){\makebox(0,0)[cl]{$\Beee$}}

\linethickness{0.35mm}
\multiput(20,70)(0.12,-0.12){167}{\line(1,0){0.12}}
\linethickness{0.35mm}
\put(20,50){\line(1,0){20}}
\linethickness{0.35mm}
\put(20,20){\line(0,1){20}}
\linethickness{0.35mm}
\multiput(0,70)(0.12,-0.12){167}{\line(1,0){0.12}}
\put(30,60){\makebox(0,0)[cc]{$\dottt$}}

\put(90,10){\makebox(0,0)[cc]{$\dottt$}}

\put(5,5){\makebox(0,0)[cc]{$\Mhh$}}

\put(120,30){\makebox(0,0)[cc]{$\Univv$}}

\linethickness{0.35mm}
\put(0,70){\line(1,0){20}}
\linethickness{0.75mm}
\put(10,10){\line(0,1){50}}
\linethickness{0.35mm}
\multiput(0,20)(0.12,-0.12){167}{\line(1,0){0.12}}
\linethickness{0.35mm}
\put(0,0){\line(0,1){20}}
\linethickness{0.35mm}
\put(0,0){\line(1,0){20}}
\put(10,10){\makebox(0,0)[cc]{$\dottt$}}

\put(20,60){\makebox(0,0)[cc]{$\prtls$}}

\linethickness{0.35mm}
\put(180,40){\line(1,0){40}}
\linethickness{0.35mm}
\put(220,20){\line(0,1){20}}
\linethickness{0.75mm}
\put(190,40){\line(0,1){40}}
\linethickness{0.35mm}
\put(200,20){\line(1,0){20}}
\put(150,30){\makebox(0,0)[cc]{\EQLS}}

\linethickness{0.35mm}
\multiput(160,30)(0.12,-0.12){167}{\line(1,0){0.12}}
\linethickness{0.35mm}
\put(210,0){\line(0,1){20}}
\linethickness{0.35mm}
\put(210,40){\line(0,1){40}}
\put(193.75,80){\makebox(0,0)[cl]{$\PPh$}}

\put(187.5,0){\makebox(0,0)[cr]{$\XH$}}

\linethickness{0.35mm}
\put(190,0){\line(0,1){10}}
\linethickness{0.75mm}
\put(170,10){\line(0,1){10}}
\put(165,5){\makebox(0,0)[cc]{$\Mhh$}}

\put(207.5,0){\makebox(0,0)[cr]{$\Aee$}}

\put(213.75,80){\makebox(0,0)[cl]{$\Beee$}}

\linethickness{0.35mm}
\multiput(160,20)(0.12,-0.12){167}{\line(1,0){0.12}}
\linethickness{0.35mm}
\put(160,0){\line(1,0){20}}
\linethickness{0.35mm}
\put(160,0){\line(0,1){20}}
\put(170,10){\makebox(0,0)[cc]{$\dottt$}}

\put(205,30){\makebox(0,0)[cc]{$\Univv$}}

\linethickness{0.75mm}
\put(190,20){\line(0,1){10}}
\linethickness{0.35mm}
\put(180,10){\line(1,0){20}}
\linethickness{0.35mm}
\put(160,30){\line(1,0){20}}
\linethickness{0.35mm}
\multiput(180,30)(0.12,-0.12){167}{\line(1,0){0.12}}
\linethickness{0.35mm}
\multiput(180,40)(0.12,-0.12){167}{\line(1,0){0.12}}
\put(190,20){\makebox(0,0)[cc]{$\dottt$}}

\put(180,20){\makebox(0,0)[cc]{$\prtls$}}

\end{picture}

\end{split}
\eeq
\epr

\section{From natural science to artificial delusions}\label{Sec:Outro}
% !TEX root = 00-wollic.tex

\subsection{What did we learn?}
We sketched the category $\UUU$ of state spaces $A, B,\ldots$, comprised of theories with reference models.  A transition $f\colon A\to B$ transforms $A$-states to $B$-states. Such morphisms capture theory expansions, reinterpretations, and map observables of type $A$ to observables of type $B$. They can be construed in terms of dynamic logic and support reasoning about the evolution of software systems or scientific theories. The crucial point is that the category $\UUU$ contains a universal language $\DP$ of explanations and belief updates. The self-reference in such languages was the crux of G\"odel's incompleteness constructions. While G\"odel established that \emph{static}\/ theories capable of self-reference cannot be complete or prove their own consistency, we note that \emph{dynamic}\/ theory and model updates allow constructing testable theories that preempt falsification. While a static model of a given theory fixes a space of true statements once and for all, the availability of dynamic semantical updates opens up the floodgates of changing models and varying notions of truth. Faster learners conquer this space faster. The bots, as the fastest learners among us, have been said to acquire their delusions from our training sets. The presented constructions suggest that they may also become delusional by dynamically updating their belief states and steering their current explanations of reality into persistent consistency, resilient to further learning. They may also combine the empiric delusions from our training sets with the logical delusions  constructed in a universal language, leverage one against the other, and get the best of both worlds. 

But why would they do that?

\subsection{Beyond true and false}
Why did the Witches tell Macbeth that it is his destiny to be king thereafter, whereupon he proceeded to kill the King? Why did the Social Network have to convince its very first users that more than half of their friends were already users? Some statements only ever become true if they are announced to be true when they are false. They are self-fulfilling prophecies. There are also self-defeating claims. In the dynamic logic of social interactions, most claims interfere with their own truth values in one way or another. If I convince enough people that I am rich, I stand a better chance to become rich. If we  convince enough people that this research direction is promising and well-funded, it will become well-funded and promising. Just like true statements about nature help us to build machines and get ahead in the universe, the manipulations of truth help us get ahead in society. They are the high-level patterns of language that used to be studied in early logic right after the low-level patterns of meaning (that used to be called \emph{``categories''}). If you train a bot to speak correctly, it will start speaking convincingly as soon as it learns long enough $n$-grams. It will lie not only the static lies contained in its training set but also the lies generated dynamically, according to the rules of rational interaction.  Rhetorics used to be studied right after grammar, sophistic argumentation after syllogisms, witchcraft arose from cooking, magic from tool building. The bot religions arise along that well-trodden path.

We presented two constructions. One produces self-confirming explanations. The other one explains all future states, so it is testable but not falsifiable. Science requires that its theories are testable and falsifiable. Religion explains all future observations. If you train a bot on long enough $n$-grams, it may arrive at persistently unfalsifiable false beliefs. %With slightly more work, the presented constructions can be refined to yield beliefs that steer their theories that cannot be disproved but steer their models. 

Truth be told, all of the constructions presented in this extended abstract have only been tested on toy examples. We may be just toying with logic. Nevertheless, the fact that semantical assignments are \emph{programmable}, tacitly established  by G\"odel and mostly ignored as an elephant in the room of logic ever since, seems to call for attention, as beliefs transition beyond the human carriers.

%An interested reader will probably jump to this section from wherever they might gave up trying to understand the world as a monoidal category. (A less interested reader will from a similar place probably jumped out of the paper.) The authors' decision to try to squeeze the category $\UUU$ into a 12-page paper is dubious. Trying to avoid it would make the paper into science fiction. Maybe there is a way to present the results in a simpler framework. We'll try again.

%\bibliographystyle{abbrv}
\bibliography{CT,induction,logic,modal,networks,PavlovicD,semantics,statistics}
\bibliographystyle{plain}

%\appendix
%\addcontentsline{toc}{part}{\appendixname}
%\label{Sec:Appendix}
%\input{9-cogsci-appendix}
%

\end{document}